\documentclass[final]{elsarticle}

\usepackage[left= 3cm, right=3cm]{geometry}
\usepackage{microtype}
\usepackage{graphicx}
\usepackage{booktabs} % for professional tables
\usepackage{tipa}
\usepackage{upgreek}
\usepackage{subfigure}
\usepackage{lipsum}
\usepackage{bbm}
\usepackage{mathalpha}
\usepackage{ragged2e}
\usepackage{xcolor}

\usepackage{amsmath,amssymb}
\usepackage{wrapfig}
\usepackage{dsfont}
\usepackage{enumerate}
\usepackage{graphicx}      % include this line if your document contains figures
\usepackage{natbib}        % required for bibliography
\usepackage{comment}
\usepackage{amsthm}

\newtheorem{theorem}{Theorem}
\newtheorem{lemma}[theorem]{Lemma}
\usepackage{ulem}

\DeclareMathAlphabet{\mathpzc}{OT1}{pzc}{m}{it}
\DeclareMathOperator*{\argmax}{arg\,max}
\DeclareMathOperator*{\argmin}{arg\,min}
\usepackage{bm}
\usepackage[ruled,vlined]{algorithm2e}

\begin{document}
\begin{frontmatter}%

\title{Safe Chance Constrained Reinforcement Learning for Batch Process Control}

 \author[a]{M. Mowbray}
 \author[b]{P. Petsagkourakis}
 \author[c]{E.A. del Rio-Chanona}
 \author[a]{D. Zhang \corref{cor1}} 
 \ead{dongda.zhang@manchester.ac.uk} 
 \cortext[cor1]{Corresponding author}

 \address[a]{Centre for Process Integration, School of Chemical Engineering and Analytical Science, The University of Manchester, Manchester, M13 9PL, United Kingdom}
 \address[b]{Centre for Process Systems Engineering (CPSE), Department of Chemical Engineering, University College London, Torrington Place, London, WC1E 7JE, United Kingdom}
 \address[c]{Centre for Process Systems Engineering (CPSE), Department of Chemical Engineering, Imperial College London, London, SW7 2AZ, United Kingdom}

\begin{abstract}                % Abstract of not more than 250 words.
Reinforcement Learning (RL) controllers have generated excitement within the control community. The primary advantage of RL controllers relative to existing methods is their ability to optimize uncertain systems independently of explicit assumption of process uncertainty. Recent focus on engineering applications has been directed towards the development of safe RL controllers. Previous works have proposed approaches to account for constraint satisfaction through constraint tightening from the domain of stochastic model predictive control. Here, we extend these approaches to account for plant-model mismatch. Specifically, we propose a data-driven approach that utilizes Gaussian processes for the offline simulation model and use the associated posterior uncertainty prediction to account for joint chance constraints and  plant-model mismatch. The method is benchmarked against nonlinear model predictive control via case studies. The results demonstrate the ability of the methodology to account for process uncertainty, enabling satisfaction of joint chance constraints even in the presence of plant-model mismatch.

\end{abstract}
\end{frontmatter}

\section{Introduction}
% (Titles of)/ these sections will change
Recently, there has been growing interest amongst the research community and industry in the development of reinforcement learning (RL) based control schemes \cite{shin2019reinforcement}. This is underpinned by the ability of RL to naturally account for process stochasticity and handle nonlinear dynamics, and reflected by a growing literature that demonstrates application empirically in applications ranging from set point control \cite{mowbray2021using, spielberg2019toward}, online optimisation and control of batch processes \cite{kim2020model,joshi2021application}, real time optimisation \cite{powell2020real} and production scheduling \cite{hubbs2020deep}. All of these works rely on offline simulation of a process model, with results often validated on the same model that the RL policy was trained. This implicitly considers that the model used offline is in fact a perfect description of the real process and, in the context of control, provokes the question: \textit{"if a model is available, why not use model predictive control (MPC)?"}. 
In practice, the real system is never perfectly described by the available model. 
In the presence of uncertainty, the predictions from a model may not have closed-form expression, e.g. propagation of uncertainty using Bayesian inference. 
Here lies the real attraction of RL controllers - the ability to find an optimal control policy \cite{kirk2004optimal, bertsekas1995dynamic} independently of closed-form expressions of the uncertain process dynamics, as is required by conventional finite dimensional optimization approaches such as stochastic, tube and distributionally robust MPC \cite{kouvaritakis2016model, langson2004robust,lu2020soft}. Additionally, the use of RL allows for a greater diversity of models i.e. they are not required to be smooth. 
% , and instead learn one from data \cite{spielberg2019toward}.

However, there is a dualism implicit to RL. RL is very data expensive because knowledge about the uncertain dynamics and the quality of a control policy is instead gained by sampling \cite{sutton2018reinforcement}.  Offline learning (simulation) is absolutely required due to the cost of real world data and the operational and safety risk associated with conducting the RL process online. As a result, there remains a dependence on the availability of a description of the physical system for offline simulation, which provides means to conduct preliminary learning before deployment to the real system. Despite this, few works consider the transfer of the policy \cite{petsagkourakis2019reinforcement} to the real online system, which promotes concerns for operational safety\footnote[1]{This is also placed in the scope of a wider concern regarding the \textit{interpretability} of machine learning systems}. For example, if model-process mismatch exists, constraints may be violated or the process driven to unsafe operating regimes. Given the acknowledgement that no model is a perfect description of the real process - the development of methods should consider that RL exploits the mathematical nature of the offline model. Similar concerns are addressed in \cite{hullen2020managing}.

Broadly, there are two approaches to synthesising the type of safe controller required: modifications could be made to the reinforcement learning process \cite{kumar2020conservative, agarwal2020optimistic, yu2021combo}, or modifications made to the offline model \cite{kidambi2021morel, yu2020mopo}, which can then be integrated into the RL objective. Recent works are discussed in the following with consideration directed to both operational and safety concerns.

\subsection{Safe Reinforcement Learning}
One of the earliest works in process systems engineering (PSE), which considers the online operational safety of reinforcement learning is provided by \cite{lee2005approximate}. Here, the authors present an action-value method, with integration of a Parzen probability density estimator \cite{parzen1962estimation} to bias the action-value function approximation based on the local data density. In this case, the data is used to construct the action-value function and hence the data density helps quantify epistemic uncertainty (i.e. the reducible part of model uncertainty arising from a lack of information - data or knowledge - about the underlying functional \cite{hullermeier2019aleatoric}). This concept is shared in more recent work \cite{clements2020estimating}, and enables the implementation to produce conservative controls and restricts optimization from exploiting the mathematical nature of the approximate action-value function. However, this approach does not consider operational constraints or the accuracy of the underlying model. For RL to be deployed to real process systems, operational constraints should be satisfied with high probability (if soft). One approach to achieve this is underpinned by modification of the control selected by the RL agent, in order to ensure the system remains within some safe set via direct optimal control (DOC) \cite{li2021safe, wabersich2021predictive}. However, the use of DOC retains explicit dependence upon a process model and imposes non-trivial learning rules that could affect the optimality of the policy produced. %Similarly, in \cite{choi2020reinforcement}, the authors propose to integrate control barrier functions (CBFs) and control lyapunov functions (CLFs), with RL used to determine the parameters describing uncertainty in the CLF and CBF rather than direct control selection. This method is therefore still reliant upon a description of the nominal process dynamics. 

Other methods directly leverage the Markov decision process (MDP) formulation, upon which the reinforcement learning problem is built. This approach tends to avoid DOC and promotes use of 'model-free' methods. A reasonably popular approach to address constraints in the RL setting is provided by the constrained MDP (CMDP) formulation. In \cite{achiam2017constrained}, the authors approach the need for satisfaction of operational constraints via CMDP, but do so in expectation and simultaneously negate process-model mismatch. In \cite{huh2020safe}, the authors propose the identification of a lyapunov function (this time model-free and outside of the CMDP framework) to ensure the process stays within some safe set with a given probability. However, potential issues arising from plant-model mismatch are similarly ignored in offline simulations. In \cite{leurent2020robustadaptive}, an approach to robust control is presented (i.e. the method optimizes for the worst case event), and the presence of process-model mismatch is considered. However, the framework is limited to linear systems with additive uncertainty. Recently, in \cite{peng2021separated} the authors present an approach to address high probability constraint satisfaction based on the augmented lagrangian. However, the penalty term presented does not provide information about the quality of control selection (i.e. essentially ignoring the RL problem) and is likely to lead to conservative control policies. There have been two methods proposed recently, by \cite{petsagkourakis2020chance,pan2020constrained}, which integrate a similar penalty method into the RL problem properly, and achieve high probability constraint satisfaction. This is achieved through deployment of the concept of constraint tightening, which is common to the stochastic MPC (sMPC) community \cite{mesbah2019stochastic, valdez2019novel, rafiei2020integration}. A further method has been proposed by \cite{yoo2021dynamic} for the case of hard constraints, which constructs a slow non-stationary MDP to promote stability of learning via the implementation of a dynamic penalty method. However, the aforementioned works negate the presence of offline model-process mismatch.

Most of the previous works ignore issues arising from process-model mismatch. The domain of batch RL (otherwise known as offline RL) has drawn a lot of recent research interest \cite{Neurips20OfflineRL}. The promise of this field lies in the synthesis of real-world control policies from existing datasets (offline). The key idea in batch RL is to learn with awareness of the limitations of the available data. Many of the works set in this domain focus on action-value methods and look to bias (or regularise) the action-value function approximation \cite{agarwal2020optimistic} by considering the data density \cite{kumar2020conservative} in a manner not dissimilar to \cite{lee2005approximate}. More recently, attention has been directed towards considerate construction of an offline model, based on the available data and this directs attention in the following analysis.

\subsection{Uncertainty Aware Modelling and Control}

A key consideration in the development of model-based RL approaches is the relationship between model construction and policy learning. For example, in \cite{rajeswaran2020game}, the problem of learning under the limitations of a local model and improving policy performance on the real process is considered within a game theoretic framework (similar to model-based design of experiments). However, it is not clear as to whether this approach would ensure real-process safety unless modifications were made to the reward function. This problem is approached by the work presented in \cite{petsagkourakis2020safe} and more recently in \cite{kidambi2021morel}. In \cite{zanon2020safe}, the authors integrate RL into a robust, linear MPC scheme, by using an RL policy to parameterise an uncertainty set. This allows for ensurance of \textit{optimality} under the scheme, but is traded at the price of restrictive modelling assumptions. In \cite{lutjens2019safe}, model uncertainty is incorporated into a penalty function for RL, however, the uncertainty estimate is gained through approximate methods such as bootstrapping and MC dropout, which provides computational cost. In \cite{kidambi2021morel}, the epistemic uncertainty associated with offline prediction is quantified via the variance of a model ensemble. The epistemic uncertainty is used to modify the reward function of the MDP to synthesise a safe control policy without further interaction with the real system. A type of model, which achieves this more naturally than an ensemble, is the Gaussian process (GP). GPs are data-driven models and their use is well documented in PSE applications \cite{sternberg2017identification,frigola2015bayesian, bradford2020stochastic, bradford2018dynamic, del2021real}. In part, this is due to their compatibility with small datasets, but primarily for their natural quantification of epistemic and aleatoric uncertainty. In a number of previous works, (realisations of) GPs have been used to inform control decisions. Most of these works lie in the domain of sMPC \cite{bradford2020stochastic, umlauft2018scenario}, however, a few hail from the field of RL-based policy optimization \cite{deisenroth2011pilco, curi2020efficient, berkenkamp2017safe}. In \cite{deisenroth2011pilco}, the authors compute gradients for policy improvement analytically, resulting in a highly efficient algorithm for unconstrained problems. In \cite{curi2020efficient}, the authors utilise GPs and the variance of the posterior distribution to produce a controller-directed exploration strategy, but negate propagation of model uncertainty and, again, process constraints. Whereas \cite{berkenkamp2017safe} present an algorithm that simultaneously balances exploration and exploitation of a GP model, considers constrained problems and provides stability guarantees for the policy identified. In the following, we draw from works closer to sMPC \cite{umlauft2018scenario, bradford2020stochastic}, to synthesise a safe RL-based control policy, which considers both operational constraints and process-model mismatch.

\subsection{Contribution}\label{sec:cont}

A number of RL-based methodologies have been proposed to ensure operational constraints are satisfied with high probability \cite{pan2020constrained, petsagkourakis2020chance, peng2021separated}. Other works have been proposed to consider the process-model mismatch that exists when learning an RL policy offline \cite{kumar2020conservative,kidambi2021morel, yu2021combo}. However, as far as the authors are aware, there are no RL methods, which achieve both. In this work, we propose a method that synchronously satisfies operational constraints with high probability, whilst respecting the limitations of a process model. Specifically, we deploy the use of GPs to construct a data-driven state space model. The variance of the posterior predictive distribution of the GP is used in two different ways: firstly, it provides a constraint tightening mechanism to back the nominal (or expected) process away from the constraint boundary (to provide constraint satisfaction with high probability); and, secondly, it is used to penalise exploration of regions of the GP model with high epistemic uncertainties. The full method as proposed also implements a Bayesian optimization strategy in order to tune the degree of constraint tightening - balancing operational risk with performance. Here, we draw analogue to reward shaping, except in this case, we identify a policy variant mechanism for constraint satisfaction as desired \cite{ng1999policy}. Importantly, the dimensionality of the shaping problem is equivalent to the number of operational constraints imposed on the system, which provides means to scale the method to larger problems. Further advantages include the inheritance of the MDP framework - which theoretically enables us to account for uncertainty in a proper closed loop manner - as well as the mitigation of resolving an optimization problem online (as is required by conventional methods). Instead controls are selected via inference, which lends itself naturally to handling systems of both fast and slow dynamics. Additionally, the approach is completely data-driven and synchronously accounts for model uncertainty, removing demands for assumption of mechanistic process knowledge.

The following is structured as follows: in Section \ref{sec:ProblemS}, we outline the problem statement and implicitly define the processes of interest; in Section \ref{sec:Method}, the methodology is presented; in Section \ref{sec:CS} a fed-batch bioprocess case study is presented with a view to demonstrate the methodology; in Section \ref{sec:R&D} and \ref{sec:conc} the results and discussion, and conclusion are presented, respectively.

\section{Problem Statement}\label{sec:ProblemS}
% minimise SOCP chat
This work is concerned with the synthesis of an optimal control strategy for nonlinear, uncertain systems of the form:
\begin{equation}\label{eq:nonlinDynA}
\begin{aligned}
    \mathbf{x}_{t+1} = f(\mathbf{x}_t, \mathbf{u}_t, \mathbf{s}_t)
\end{aligned}
\end{equation}
where  $\mathbf{x} \in \mathbb{X} \subseteq \mathbb{R}^{n_x}$ denotes the system state; $\mathbf{u} \in \mathbb{U} \subseteq \mathbb{R}^{n_u}$ the control inputs to the system; $\textit{t}=[1, \ldots, T]$ denotes the discrete time index; $\mathbf{s} \in \mathbb{S} \subseteq \mathbb{R}^{n_s}$, where $\mathbb{S}$ represents a set of realisations of process stochasticity; and, $f: \mathbb{X} \times \mathbb{U} \times \mathbb{S} \rightarrow \mathbb{X}$. Here, no formal assumption is made regarding the source of stochasticity $\mathbb{S}$, but it could be introduced via parametric uncertainty or disturbances. In either case, given the presence of stochasticity within system description, Eq. \ref{eq:nonlinDynA} may be expressed equivalently via the following conditional probability density function:
\begin{equation}\label{eq:condensity}
\begin{aligned}
    \mathbf{x}_{t+1} \sim p(\mathbf{x}_{t+1}| \mathbf{x}_{t}, \mathbf{u}_t)
\end{aligned}
\end{equation}
Specifically, it is assumed that the process dynamics adhere to description as a Markov process, and therefore that the associated decision-making problem may be formalized as a Markov decision process (MDP). MDPs provide a probabilistic value framework for decision making in uncertain systems, which display the Markov property. Under the MDP framework, the probability of observing a given process trajectory $p(\bm{\tau})$, under a control policy $\pi$ is described:
\begin{equation}\label{eq:tauevolve}
    \begin{aligned}
        p(\bm{\tau}) = p(\mathbf{x}_0) \prod_{t=0}^{T-1} \pi(\mathbf{u}_t | \mathbf{x}_{t})p(\mathbf{x}_{t+1}|\mathbf{x}_t, \mathbf{u}_t)  
    \end{aligned}
\end{equation}
where $\bm{\tau} = (\mathbf{x}_0,\mathbf{u}_0, \ldots, \mathbf{x}_T)$ denotes the process trajectory; $p(\mathbf{x}_0)$ denotes the initial state distribution; $p(\mathbf{x}_{t+1}|\mathbf{x}_t, \mathbf{u}_t) $ the process dynamics; and the policy $\pi(\mathbf{u}_t|\mathbf{x}_t)$ is explicitly defined as a conditional probability function over control inputs. Provided process evolution is subject to a stochastic policy and process dynamics, the performance of a policy is evaluated via the expected discounted sum of rewards $R_{t+1} \in \mathbb{R}$ accumulated from the initial state:
\begin{equation}\label{eq:objective}
    \begin{aligned}
        G(\bm{\tau}) &= \sum_{t=0}^{T-1} \gamma^{t}R_{t+1} \\
        J &= \int p(\bm{\tau})G(\bm{\tau}) d\bm{\tau}
    \end{aligned}
\end{equation}
where the reward is allocated by a reward function $R: \mathbb{X} \times \mathbb{U} \times \mathbb{X} \rightarrow R_{t+1}$ and $\gamma = [0,1]$ is the discount factor. Therefore, the optimal policy $\pi^*$:
\begin{equation}
    \begin{aligned}
        \pi^* &= \argmax_{\pi} J
    \end{aligned}
\end{equation}
One approach to learning such a controller is via Reinforcement Learning (RL). However, under the framework provided by MDPs, the optimal policy $\pi^*$ (and, hence RL) implicitly neglects the satisfaction of both safety and operational constraints. In applications related to this work (i.e. industrial batch process systems), the satisfaction of both operational and safety constraints is of concern. As such, it is of interest to develop an RL-based methodology for the synthesis of an optimal control policy $\pi^*_C$, which respects constraints. The problem statement follows that common to works set in the domain of stochastic optimal control:
\begin{equation}
\mathcal{P}(\pi_C):=\left
\{\begin{aligned}
        &\max_{\pi} J\\
    &\text{s.t.}\\
    &\textbf{x}_0 \sim p(\textbf{x}_0)\\
    &\textbf{x}_{t+1} \sim p(\textbf{x}_{t+1}|\textbf{x}_t, \textbf{u}_t)\\
    &\textbf{u}_t \sim \pi(\textbf{u}_t|\textbf{x}_t)\\
    &\textbf{u}_t\in\mathbb{\hat{U}}\\
    &\mathbb{P}(\bigcap_{i=0}^T \{\textbf{x}_i \in \mathbb{\hat{X}}_i\})\geq 1-\alpha\\
    &  \forall t \in \left\{0,...,T-1\right\}\label{eq:OCP}
\end{aligned}\right.
\end{equation}
where $\mathbb{\hat{U}}\subset \mathbb{U}$ represents the set of control inputs, which satisfy hard constraints on the control space; and, $\mathbb{\hat{X}}  \subset \mathbb{X}$ denotes the set of states, which satisfy operational and safety constraints imposed on the state space. Under the assumption that the problem definition may have $n_g$ constraints, $\mathbb{\hat{X}}$ may be expanded more generally as the \textit{joint} chance constraint set, such that:
\begin{equation}\label{eq:safeset}
    \begin{aligned}
    \mathbb{\hat{X}}_t = \{ \mathbf{x}_t \in \mathbb{G}_{j,t}, \forall j \in \{1, \ldots, n_g\}\}
    \end{aligned}
\end{equation}
where $\mathbb{G}_{j,t} \subset \mathbb{R}^{n_x}$ defines the set of states, which ensure satisfaction of the $j^{th}$ constraint at time step $t$. Specifically, in the following analysis, we assume that:
\begin{equation}
    \begin{aligned}
        \mathbb{G}_{j,t} = \{\mathbf{x}_{t} \in \mathbb{R}^{n_x} : A_j^T\mathbf{x}_{t} - b_j \leq 0\} 
    \end{aligned}
\end{equation}
where $A_j \in \mathbb{R}^{n_x}$ and $b_j\in \mathbb{R}$ define the $j^{th}$ constraint. The general principles discussed subsequently extend to problems with nonlinear constraints. However, in that case, the constraints should be represented by lower order power series expansions of the nonlinear functions \cite{rafiei2018stochastic} i.e. the nonlinear expressions should be linearized. Given that the process is stochastic, the constraints are 'softened' such that satisfaction is guaranteed for all time $t=\{0,\ldots,T\}$ with a desired probability, denoted $1-\alpha$.  

Theoretically, solution to Eq. \ref{eq:OCP} may be realised via exact dynamic programming (DP), which requires exact descriptions of the probabilistic process dynamics. In process systems, these are typically unavailable. Further, DP is known to suffer from the \textit{the curse of dimensionality}, which implies that high dimensional problems, or those that operate over continuous state and control spaces, are computationally intractable. In the domain of sMPC, works generally leverage reformulation of the problem via deterministic expressions for the joint chance constraints and modelling assumptions regarding the nature of process stochasticity \cite{mesbah2019stochastic, Subramanian_2021}. This work similarly forms a deterministic surrogate of Eq. \ref{eq:OCP} in combination with Gaussian process (GP) data-driven modelling, and identifies a reinforcement learning (RL) based control policy, which naturally accounts for process stochasticity in a closed-loop manner. These benefits are complementary to those noted in Section \ref{sec:cont}. In the following section, a methodology is proposed for synthesis of the controller $\pi_C$. 

\section{Methodology}\label{sec:Method}

\subsection{Gaussian Processes for Data-Driven Dynamic Modelling}\label{sec:GPdyna}

Model-free RL-based policies are learned through Monte Carlo (MC) sampling of the process dynamics and iteratively improved based on the collected data. This is otherwise known as policy iteration. For real world applications, the synthesis of RL-policies is dependent upon an accurate description (model) of the process dynamics. For nonlinear, uncertain processes, construction of mechanistic dynamical models can be problematic, even if understanding of the fundamental mechanisms driving process behaviour exists. Hence, the construction of a purely data-driven model is proposed to represent the discrete time, evolution of the nonlinear, uncertain dynamical system described by $\mathbf{x}_{t+1} = f(\mathbf{x}_t, \mathbf{u}_t, \mathbf{s}_t)$, i.e. Eq \ref{eq:nonlinDynA}. In order to construct a representation of the system dynamics, it is assumed that: a) $f$ is a smooth function and b) there is an available dataset $\mathcal{D}$, which is composed as follows:
\begin{equation}\label{eq:Dataset}
\begin{aligned}
    \mathcal{D} = [\bm{\Upsilon}^T\ \mathbf{Y}^T ], \qquad \mathbf{Y} = [\mathbf{y}_i, \ldots, \mathbf{y}_N],\qquad
    \bm{\Upsilon} = [\bm{\upsilon}_i, \ldots, \bm{\upsilon}_N], \qquad
    \bm{\upsilon}_i = \big[ \mathbf{x}_i^T \ \mathbf{u}_i^T \big]^T, 
\end{aligned}
\end{equation}
where $\bm{\upsilon} \in \mathbb{V} \subseteq \mathbb{R}^{n_{\upsilon}}$, $n_{\upsilon}=n_x+n_u$ are input measurements and $\mathbf{y} \in \mathbb{Y} \subseteq \mathbb{R}^{n_x}$ are output measurements of the system, which are gathered subject to some noisy process $\bm{\omega} \in \mathbb{W} \subseteq \mathbb{R}^{n_x}$ \cite{Rasmussen06gaussianprocesses}. Here, $\mathbb{W}$ is assumed to be an infinite set representative of possible realisations of system noise, such that:
\begin{equation}\label{eq:noisygen}
\begin{aligned}
    \mathbf{y}_i    &= f(\bm{\upsilon}_i) + \bm{\omega}_i \\
    \bm{\omega}_i          &\sim \mathcal{N}(0, \Sigma_n)
\end{aligned}
\end{equation}
where $\Sigma_n = diag([\sigma^2_{n,1}, \ldots, \sigma^2_{n,n_x}]) \in \mathbb{R}^{n_x \times n_x}$ defines a diagonal matrix, where each element on the diagonal denotes a state dependent variance. Further, as usual, it is assumed that all datapoints $d_i = [\bm{\upsilon}_i , \mathbf{y}_i] $ (equivalent to rows of $\mathcal{D}$) are independently and identically distributed (i.i.d.). Parallel can be drawn between Eq. \ref{eq:condensity}, such that Eq. \ref{eq:noisygen} is equivalently described as a conditional probability function $\mathbf{y} \sim p(\mathbf{y}|\bm{\upsilon})$. This description of data generation shares similarities to assumptions made in Section \ref{sec:ProblemS} and directs attention to a branch of probability theory known as stochastic processes (SPs), and in particular Gaussian processes (GPs).

\subsubsection{Gaussian Processes}\label{sec:GP}

SPs define a probability model over an infinite collection of random variables, any finite subset of which have a joint distribution \cite{lindgren2012stationary}. This definition leads to the interpretation of SPs as probability distributions over functions \cite{Rasmussen06gaussianprocesses}, such that one realisation of an SP can be thought of as obtaining a sample from a function space. When the distribution over the function space is assumed Gaussian, the resultant model is termed a GP. 

A GP is fully specified by a mean function, $\mathbf{m}:\mathbb{V} \rightarrow \mathbb{R}$, and covariance function, $k:\mathbb{V} \times \mathbb{V} \rightarrow \mathbb{R}$, such that:
\begin{equation}\label{GPapprox}
\begin{aligned}
    f_{GP}(\bm{\upsilon}) &\sim \mathcal{GP}\big(\mathbf{m}(\bm{\upsilon}), k(\bm{\upsilon}, \bm{\upsilon}{'})\big)
\end{aligned}
\end{equation}
A number of covariance functions exist within the GP toolbox. Selection of both the function and the associated hyperparameters, $\bm{\lambda}\in \mathbb{R}^{n_\lambda}$, define the properties of the GP in function space. As such, the decision as to appropriate covariance function is often informed by domain knowledge and understanding of the modelling problem at hand. The definition of hyperparameters is handled by maximisation of the marginal log-likelihood (this is discussed in \ref{sec:GPtrain} and referred to as GP training). Popular choices include the Matern 5/2 and radial basis function (RBF) covariance functions \cite{Rasmussen06gaussianprocesses}. Definition of the mean function is also important.  Often, a zero mean ($\mathbf{m}(\bm{\upsilon}) = 0$) is assumed, which is not unreasonable given standardisation of the output data, $\mathbf{Y}$.

GP model inference takes place within the framework provided by Bayesian reasoning. The assertion of a modelling decision regarding the mean and covariance function therefore represents a prior belief about the possible properties of the hidden, functional relationship expressed in the dataset $\mathcal{D}$. When presented with a new test input $\bm{\upsilon}^* \in \mathbb{R}^{n_{\upsilon}}$, the construction of a single GP model for the $j^{\textit{th}}$ state leads to the generation of an associated prediction $y_j^* \in \mathbb{R}$ via the following joint prior distribution:
\begin{equation}\label{eq:JointDist}
\Bigg[\begin{aligned}
    \begin{matrix} \mathbf{Y}_j^T \\ y_j^* \end{matrix} \Bigg] &= \mathcal{N} \Bigg(0, \Bigg[\begin{matrix} K + \sigma_{n}^2I_N & K_* \\
    K_*^T & k(\bm{\upsilon}^*, \bm{\upsilon}^*) \end{matrix} \Bigg] \Bigg) 
\end{aligned}
\end{equation}
where $\mathbf{Y}_j \in \mathbb{R}^{1\times N}$ denotes the $j^{\textit{th}}$ row of the output of the training dataset $\mathbf{Y}$; $K\in \mathbb{R}^{N \times N}$ denotes the Gram matrix, such that provided with training input measurements (see Eq. \ref{eq:Dataset}),  element $k_{m,n} = k(\bm{\upsilon}_m, \bm{\upsilon}_n)$, where $m = [1, \ldots, N]$ and $n = [1, \ldots, N]$; $\sigma^2_{n}$ denotes the variance of the noise associated with observation of state $y_j \in \mathbb{R}$ (see Eq. \ref{eq:noisygen}); $K_* \in \mathbb{R}^{N}$ denotes the covariance of the test datapoint $\bm{\upsilon}^*$ with the existing (training) input measurements; and, lastly, $k(\bm{\upsilon}^*, \bm{\upsilon}^*) \in \mathbb{R}$ represents the variance of the test datapoint. 

Furthermore, as GPs operate through Bayesian reasoning, by \textit{conditioning} the joint prior distribution (Eq. \ref{eq:JointDist}) upon the observed dataset $\mathcal{D}$  and the test point $\bm{\upsilon}^*$, we obtain a predictive posterior Gaussian distribution, with mean $\mu_j$ and variance $\sigma^2_j$ as follows:
\begin{equation}\label{eq:posterior}
\begin{aligned}
\mu_j(\bm{\upsilon}^*) &= K_*^T(K + \sigma_{n}^2I_N)^{-1}\mathbf{Y}_j^T \\
\sigma^2_j(\bm{\upsilon}^*) &= k(\bm{\upsilon}^*, \bm{\upsilon}^*) - K_*^T(K + \sigma_{n}^2I_N)^{-1}K_*
\end{aligned}
\end{equation}
In the context of dynamical systems modelling, Eq. \ref{eq:posterior} represents a probability model over the next state of the dynamical system at the next discrete time index.  The construction of a posterior probability function is particularly useful in engineering applications, given that it expresses elements of both aleotoric and epsitemic model uncertainty. Typically, the mean is taken as the model's prediction, however, prediction may also be directly sampled from posterior distribution \cite{bradford2018dynamic}. This will be discussed further in section \ref{sec:GPReal}. 

Thus far, the methodology has formalised the construction of GPs, and defined them as multiple-input, single-output models. Hence a single GP provides a functional mapping descriptive of the future discrete time evolution of a single state, given observation of the full system state and control inputs at the current time index. It is of interest to this work to construct a multiple-input, multiple-output state space model. This is discussed subsequently in Section \ref{sec:GPSS} and has been presented previously by other related works \cite{bradford2020stochastic, umlauft2018scenario, deisenroth2011pilco}. We direct the interested reader for more information. 

\subsubsection{Gaussian Processes for State Space Modelling}\label{sec:GPSS}

In this study, state space models are constructed by training $n_x$ GP models separately and combining them to simultaneously predict the state vector $\mathbf{x} \in \mathbb{R}^{n_x}$ at the next discrete time interval, $t+1$. Specifically, under the assumption that each of the $n_x$ models has been constructed and trained according to Section \ref{sec:GP} and \ref{sec:GPtrain}, this implies that the the posterior prediction from the GP state space model, when presented with $\bm{\upsilon}_t$ follows:
\begin{equation}\label{eq:GPSS}
    \begin{aligned}
    \bm{\mu}(\bm{\upsilon}_t; \mathcal{D}) &= \big[\mu_1(\bm{\upsilon}_t), \ldots, \mu_{n_x}(\bm{\upsilon}_t)\big] \\
    \bm{\Sigma}(\bm{\upsilon}_t; \mathcal{D}) &= diag(\sigma^2_1(\bm{\upsilon}_t), \ldots, \sigma^2_{n_x}(\bm{\upsilon}_t)) \\
    \mathbf{x}_{t+1} &\sim \mathcal{N}\big(\bm{\mu}, \bm{\Sigma} \big)
    \end{aligned}
\end{equation}
where $\bm{\mu} \in \mathbb{R}^{n_x}$ and $\bm{\Sigma} \in \mathbb{R}^{n_x \times n_x}$ amd $\mathbf{x}_{t+1} \in \mathbb{R}^{n_x}$ is the next state . In the following section, we discuss how the GP state space model is used to generate realisations of underlying process stochasticity, and relate discussion directly to the decision making process.  

\subsubsection{Gaussian Process Realisations and Decision Making}\label{sec:GPReal}

For effective and safe control and optimization of process systems, a control policy must consider worst-case realisations of process stochasticity. In GP models, function realisations are sampled from the GP. Each function realisation represents a specific instance of model uncertainty across process evolution - including the worst case. In order to achieve this, model uncertainties must be propagated correctly. This work implements the method detailed in \cite{umlauft2018scenario}, which recursively updates the dataset $\mathcal{D}$ as the process evolves between discrete time indices. This process is detailed by Algorithm \ref{Alg:UncertProp} in \ref{sec:MCGP} and relies upon linear algebra to account for the effects of conditioning the GP models on the updated dataset. See \cite{bradford2020stochastic, strassen1969gaussian} for more details.

In the following section, an approach that synchronously combines concepts from sMPC and RL to produce a self-optimizing, policy varying reward shaping mechanism is presented, which provides probabilistic constraint satisfaction. Specifically, a penalty function method is combined with the concept of backoffs.
%
%
%, such that after observing the first discrete increment $\mathbf{y}_0$ from a given initial state and control pair $\bm{\upsilon}_0 = \big[ (\mathbf{x}_0)^T, (\mathbf{u}_0)^T \big]^T$ the dataset is described $\mathcal{D}_0 = (\mathcal{D}, (\bm{\upsilon}_0, \mathbf{y}_0))$. Here, $\mathbf{y}_0$ is sampled from the posterior normal distribution and $\mathbf{u}_0$ is dictated by the decision making element. When constructing the posterior distribution to determine the second discrete increment $\mathbf{y}_1$, we condition the GP on the updated dataset $\mathcal{D}_0$ and the new testpoint $\bm{\upsilon}_1 = \big[ (\mathbf{y}_0)^T, (\mathbf{u}_1)^T \big]^T$, where again, $\mathbf{u}_1$ is derived from the decision-maker. This process continues for the full control horizon. More generally, this recursive updated dataset may be written $\mathcal{D}_+ = (\mathcal{D}, (\bm{\upsilon}_+, \mathbf{y}_+))$ 
%

\subsection{Safe Chance Constrained Policy Optimization with Gaussian Processes}

In this section, we provide details of the methodology, which enables combination of GP state space models with RL-based policy optimization for high probability constraint satisfaction. To achieve this, the methodology is organised as follows and the full algorithm is detailed by Algorithm \ref{Alg:BOBackoffs}:
\begin{enumerate}
    \item In Section \ref{sec:DeterSurr}, the general stochastic optimal control problem defined by Eq. \ref{eq:OCP} is modified to consider the nominal evolution of the states and obtain a deterministic expression for the probabilistic joint constraints. To facilitate this, we implement an approach similar to \cite{petsagkourakis2020chance} in combination with a GP state space model. 
    \item In Section \ref{sec:constrainedPPO} the deterministic surrogate constraints are incorporated into a reformulation of the RL objective (see Eq. \ref{eq:objective}) via an $\textit{l}_\textit{p}$ penalty function\footnote[2]{The subscript \textit{p} of $\textit{l}_\textit{p}$ denotes the norm incorporated into the penalty function} \cite{nocedal2006numerical, larson2019derivative} and detail of a general constrained policy optimization algorithm is provided \cite{schulman2017proximal}. 
    \item Then, in Section \ref{sec:BO}, an 'efficient' global optimization strategy \cite{jones1998efficient} is presented to iteratively tune the penalty function enabling satisfaction of the original joint chance constraints with the desired probability $1-\alpha$ establishing a strong connection between this work and reward shaping \cite{ng1999policy}.
\end{enumerate}
The methodology is formalized with a view to the use of policy optimization methods, however, the concepts discussed can also be integrated into actor-critic and action-value methods \cite{sutton2018reinforcement}.

\subsubsection{Probabilistic Joint Chance Constraints}\label{sec:DeterSurr}

In this section, reformulation of the probabilistic joint chance constraint detailed by Eq.\ref{eq:OCP} is presented. The joint chance constraints are restated here for ease:
\begin{equation}\label{eq:jointChance}
    \begin{aligned}
    \mathbb{P}(\bigcap_{i=0}^T \{\textbf{x}_i \in \mathbb{\hat{X}}_i\})\geq 1-\alpha\\
    \end{aligned}
\end{equation}
The following analysis proceeds to obtain a set of deterministic surrogate constraints, which can then be integrated into a revised objective for RL-based policy optimization. In particular, we leverage Boole's inequality and the Cantelli-Chebyshev inequality to obtain a deterministic constraint for each of those that comprise the original joint constraint. The analysis follows \cite{paulson2020stochastic, farina2014mpc}.
\begin{lemma}\label{lem:Boole's}
\textbf{Boole's Inequality} \cite{boole1847mathematical}: Consider a finite set of countable events $\{Z_1, Z_2, \ldots, Z_{n_g}\}$, the probability that one of these events occurs is no greater than the sum of the probabilities of the individual events:
\begin{equation}
    \begin{aligned}
    \mathbb{P}\Big(\bigcup_{i=1}^{n_g} Z_i \Big) \leq \sum_{i=1}^{n_g} \mathbb{P}(Z_i)
    \end{aligned}
\end{equation}
\end{lemma} 
Now, considering $n_g$ constraints comprise the joint chance constraint, then applying this result enables decomposition of Eq. \ref{eq:jointChance}, into $n_g$ individual chance constraints. As in \cite{petsagkourakis2020chance}, for ease of notation, we define the following:
\begin{align*}
    {X} = \max_{(t,j) \in \{0, \ldots, T\} \times \{1, \ldots, n_g\}} A_j\mathbf{x}_{t} - b_j,  \qquad  g = \{\mathbf{x}\in \mathbb{R}^{n_x}: X \}, \qquad    \mathbb{G}^{'}_j = \bigcap_{i=0}^T \{\textbf{x}_i \notin \mathbb{G}_{j,i}\}
\end{align*}
where $\mathbb{G}^{'}_j$ defines the set of states, which do not satisfy constraint $j$ for all time indices and $X\in \mathbb{R}^{n_x}$ defines a random variable. From Lemma \ref{lem:Boole's}:
\begin{equation}\label{eq:Boole}
    \begin{aligned}
    \mathbb{P}\Big(\bigcup_{j=1}^{n_g} \{g \subset \mathbb{G}^{'}_j\} \Big) \leq \sum_{i=1}^{n_g} \mathbb{P}(g \subset \mathbb{G}^{'}_j)
    \end{aligned}
\end{equation}
Explicitly, Eq. \ref{eq:Boole}, dictates that the probability of achieving joint constraint satisfaction under a given policy $\pi$ is lower bounded by the probability of satisfying each of the respective constraints individually. Therefore, guaranteeing satisfaction of chance constraints individually can be considered a robust approximation to joint satisfaction:
\begin{align*}
\iota_j = \mathbb{P}(g \subset \mathbb{G}^{'}_j) \implies \alpha \leq \sum_{j=1}^{n_g} \iota_j
\end{align*}
where $\iota_j \in \mathbb{R}$, subject to satisfying Eq. \ref{eq:Boole}. This enables approximation of Eq. \ref{eq:jointChance} via the following:
\begin{equation}\label{eq:B2}
    \begin{aligned}
        &\sum_{j=1}^{n_g}\mathbb{P}\big(\bigcap_{i=0}^T \{\mathbf{x}_i \in \mathbb{G}_{j,i}\}\big) = 1 - \sum_{j=1}^{n_g}\iota_j
    \end{aligned}
\end{equation}    
In this work, we define $\iota_j = \alpha/{n_g},\  j =[1,\ldots, n_g]$. Having decomposed the original joint chance constraint into a set of individual chance constraints, the methodology looks to express a set of deterministic surrogate expressions (of the original probabilistic chance constraints), which can then be incorporated into the method presented. %, which is detailed by Lemma \ref{lem:CCeq}.

To proceed, we deploy the concept of \textit{constraint tightening}, which is an approach commonly deployed within the domain of sMPC. The intuition behind constraint tightening is described as follows. The process of concern is subject to unbounded uncertainties. We consider that under a given policy $\pi$, the process will vary probabilistically within a given region of $\mathbb{\hat{X}}_t$. Specifically, one can assume that the process will vary within some  euclidean distance from the nominal or expected behaviour with a given probability. If we \textit{back} the nominal process \textit{off} from the constraint boundary then we will be able to achieve chance constraint satisfaction with the desired probability. This is underpinned by the Cantelli-Chebyshev inequality, which is described by Lemma \ref{lem:CCeq}
\begin{lemma}\label{lem:CCeq}
\textbf{Cantelli-Chebyshev Inequality} \cite{ogasawara2019multiple}: Consider a random variable $Z$, with expected value $\mathbb{E}\big[Z\big]$ and finite variance $\Sigma[Z]$, then:
\begin{align*}
    \mathbb{P}\big(Z - \mathbb{E}\big[Z\big] \geq \delta\big) \leq \frac{\Sigma\big[Z]}{\Sigma[Z] + \delta^2}
\end{align*}
\end{lemma}
%
%this the variable will vary some euclidean distance from the expected coordinate is upper bounded by a scalar function of its variance and that distance. This leads to the common assumption within the field of sMPC, that if the constraint set is \textit{tightened} then chance constraint satisfaction can be guaranteed as desired. This is known as \textit{constraint tightening}.

The mechanism of constraint tightening takes the form of a set of \textit{backoffs} $\bm{\varepsilon}_j = [\varepsilon_{j,0}, \ldots, \varepsilon_{j,T}]$, which can be conceptualised as the necessary euclidean distance from the expected or nominal state $\mathbf{\bar{x}}_t \in \mathbb{R}^{n_x}$ to the constraint boundary to guarantee chance satisfaction with a given probability (note, backoff values are specific to both the constraint and time index). As stated in Section \ref{sec:ProblemS}, the analysis provided in this work assumes affine constraints. Therefore, the tightened constraint sets follow:
\begin{equation}
    \begin{aligned}
        \mathbb{\bar{G}}_{j,t} &= \{ \mathbf{\bar{x}}_t \in \mathbb{R}^{n_x} : A_j^T\mathbf{\bar{x}}_t + \varepsilon_{j,t} - b_j \leq 0 \} \\
        \mathbb{\bar{X}}_t &= \{ \mathbf{\bar{x}}_t \in \mathbb{\bar{G}}_{j,t}, \forall j =\{1,\ldots, n_g\}\} 
    \end{aligned}
\end{equation}
The determination of the backoff values $\varepsilon_{j,t}$ is handled via the following analysis. Specifically, we work from the developments made in \cite{farina2014mpc,magni2009stochastic}, which (via Lemma \ref{lem:CCeq}) show that the Cantelli-Chebyshev approximation of the backoff set is equivalent to:
\begin{equation}
    \begin{aligned}
        \varepsilon_{j,t} = \sqrt{\frac{1-\iota_j}{\iota_j}}\sqrt{A_j^T\Sigma[\mathbf{x}_t]A_j}
    \end{aligned}
\end{equation}
where $\bm{\varepsilon}_{j,t}$ represents a robust approximation of the backoff required for individual chance constraint satisfaction with the desired probability $\iota_j$. In this work, we deploy a GP state space model to estimate both the nominal state $\mathbf{\bar{x}} \in \mathbb{R}^{n_x}$ and the variance of the state $\Sigma[\mathbf{x}_t]$, as described by Eq. \ref{eq:GPSS}, enabling construction of a deterministic expression for each of the individual chance constraints. In practice, it is well documented that use of the Cantelli-Chebyshev approximation leads to overly-conservative control policies, which operate far from the constraint boundary. In order to balance the performance of the control trajectory, with constraint satisfaction, we propose to tune $\varepsilon_{j,t}$ via a multiplying factor $\xi_j = [0,1]$ for each constraint. As such, the deterministic surrogate for each of the individual chance constraints, detailed by Eq. \ref{eq:B2}, are described:
\begin{equation}
    \begin{aligned}
        A_j^T\mathbf{\bar{x}}_t + \xi_j\sqrt{\frac{1-\iota_j}{\iota_j}}\sqrt{A_j^T\Sigma[\mathbf{x}_t]A_j} - b_j \leq 0
    \end{aligned} \label{eq:backoffCC}
\end{equation}
The approach to the tuning of the multiplying factors, $\bm{\xi} = [ \xi_1, \ldots, \xi_{n_g}]$, could be handled via Bayesian optimization (BO) or bisection method \cite{petsagkourakis2020chance, bradford2020stochastic, pan2020constrained}. This work employs a BO strategy, which is detailed by Section \ref{sec:BO}. The computational implications for this are small given the efficiency of BO.

The use of a GP to parameterise the backoff values $\bm{\varepsilon}_j$ is more efficient than the set of methods proposed previously by \cite{petsagkourakis2020chance, pan2020constrained}. In those works, initial backoff values were estimated via MC sampling and then tuned. Here, we provide a method to analytically express the backoff values via the posterior predictive distribution of the GP state space model, removing the requirement for sampling and the potential inaccuracies it brings in initialisation. This is the primary novelty of this work. 

The methodology has now obtained a set of deterministic surrogate constraints for joint chance constraint satisfaction. Identification of these expressions enables reformulation of the original problem statement $\mathcal{P}(\cdot)$ described by Eq.\ref{eq:OCP} as follows:
\begin{equation}
\mathcal{\hat{P}}(\pi_C):=\left
\{\begin{aligned}
        &\max_{\pi} J\\
    &\text{s.t.}\\
    &\textbf{x}_0 \sim p(\textbf{x}_0)\\
    &\textbf{x}_{t+1} \sim \mathcal{N}(\bm{\mu}(\bm{\upsilon}_t), \Sigma(\bm{\upsilon}_t))\\
    &\textbf{u}_t \sim \pi(\textbf{u}_t|\textbf{x}_t)\\
    &\textbf{u}_t\in\mathbb{\hat{U}}\\
    &\textbf{x}_t \in \mathbb{\bar{X}}_t\\
    &  \forall t \in \left\{0,...,T-1\right\}
    \end{aligned}\right.\label{eq:approxOCP}
\end{equation}
where $\bm{\upsilon}_t = \big[ \mathbf{x}_{t}^T \ \mathbf{u}_{t}^T \big]^T $ and solution to $\mathcal{\hat{P}}(\cdot)$ is equivalent to that of the original $\mathcal{P}(\cdot)$. Due to the presence of a GP state space model within the problem description, $\mathcal{\hat{P}}$ is a function space optimization problem. Previous works have solved this problem via  nonlinear MPC with precalculation of the backoff values, and description of the discrete time state evolution according to the mean of the GP (equivalent to the nominal process)  \cite{bradford2020stochastic}. In this work, we use RL to solve $\mathcal{\hat{P}}$ (hence the use of function realisations and Algorithm \ref{Alg:UncertProp}, detailed by \ref{sec:MCGP}) with incorporation of the deterministic surrogate constraints into a modified RL objective. This is achieved via an $\textit{l}_p$ penalty function, under a given value of the backoff multipliers, $\bm{\xi}$. Solution to this problem under the optimal backoff multipliers $\bm{\xi}^*$ is deemed equivalent to finding solution to Eq. \ref{eq:approxOCP} as discussed subsequently.

\subsubsection{Safe Constrained Policy Optimization with Fixed Backoffs}\label{sec:constrainedPPO}
As GPs express process uncertainties, they present an avenue to synthesise policies, which only exploit regions of the state space in which the model is confident of the true process behaviour i.e. where epistemic uncertainties are low. By incorporating the variance prediction, $\Sigma(\bm{\upsilon}_t)$, of the GP state space model posterior directly in the RL performance index, we force the ultimate RL policy to avoid the areas that the GP is uncertain and provide explicit mechanism to mitigate exploitation of the mathematical nature of the GP model. Hence the policy \textit{pessimistically accounts for the limitations of the data-driven model} when deployed to the real process.

Use of the $\textit{l}_1$ or $\textit{l}_2$ penalty functions is particularly appealing because of the exactness (under certain conditions) to the solution of $\mathcal{\hat{P}}$ \cite{nocedal2006numerical}. This would further preserve the approximation provided by Eq. \ref{eq:approxOCP} to Eq. \ref{eq:OCP}.  The general penalty function, $\bm{\varphi}_p:\mathbb{X} \times \mathbb{U} \times \mathbb{X} \rightarrow \mathbb{R}$ is detailed as follows:
\begin{equation}
    \begin{aligned}
        \bm{\varphi}_p(\mathbf{x},\mathbf{u}, t) = R_{t+1} -  \textbf{tr}\big(\zeta\Sigma[\bm{\upsilon}_{t}]\big) - \kappa \left\lVert[A^T\mathbf{{x}}_{t+1} + \bm{\varepsilon}_{t} - b]^-\right\rVert_p
    \end{aligned}\label{eq:PF}
\end{equation}
where $A \in \mathbb{R}^{n_x \times n_g}$ and $b \in \mathbb{R}^{n_g}$ define the set of inequality constraints; $\bm{\varepsilon}_{t} \in \mathbb{R}^{n_g}$ the set of backoff values relevant to the set of constraints at a given time index (see Eq. \ref{eq:backoffCC}); $[\mathbf{z}]^- = \max(0, \mathbf{z})$ defines an element wise operation over $\mathbf{z} \in \mathbb{R}^{n_g}$; $\left\lVert \cdot \right\rVert_p$ the general \textit{p}-norm; $R_{t+1} \in \mathbb{R}$ the rewards accumulated under the original process objective e.g. productivity maximisation in a (bio)chemical process; and, $\kappa \in \mathbb{R}$ and $\zeta \in \mathbb{R}^{n_x \times n_x}$ (a diagonal matrix) weight the penalty for constraint violation and model uncertainty, respectively - relative to $R_{t+1}$. The incorporation of model uncertainty, therefore, is represented by the term $\textbf{tr}\big(\zeta\Sigma[\bm{\upsilon}_{t}]\big)$. It is expected that in some cases there is likely to be a dependence between the uncertainty and constraint penalty terms, which may lead to over-penalisation of constraint violations. This may favour the identification of conservative policies, although this is likely to be case dependent and may be mitigated by the tuning process discussed in Section \ref{sec:BO}. Expression of the penalty function, enables redefinition of the RL objective $J(\bm{\tau})$ via $\bar{J}_C(\bm{\tau})$:
\begin{equation}\label{eq:consobjective}
    \begin{aligned}
        \bar{G}_C(\bm{\tau}) &= \sum_{t=0}^{T-1} \gamma^{t}\bm{\varphi}_p(\mathbf{x},\mathbf{u}, t) \\
        \bar{J}_C&= \int p(\bm{\tau})\bar{G}_C(\bm{\tau}) d\bm{\tau}
    \end{aligned}
\end{equation}
It is hypothesised that RL-based optimization of this new objective will synthesise a policy, which provides chance constraint satisfaction and exploits regions of the state space well characterised by the model - encouraging the learning of inherently safe control policies. Further, because the modifications are made directly to the reward function itself, the approach is compatible with any RL method. Given the GP state space model is constructed over continuous state and control variables, as usual, the RL can learn a parameterisation of the optimal constrained policy:
\begin{equation}
    \begin{aligned}
        \pi_C^*(\mathbf{u}|\mathbf{x}; \theta, \cdot) &\approx \pi_C^*(\mathbf{u}|\mathbf{x})  \\
        \pi_C^*(\cdot, \theta) &= \argmax_\theta \bar{J}_C
    \end{aligned}\label{eq:funcapprox}
\end{equation}
where $\theta \in \mathbb{R}^{n_\theta} $ denotes a vector representation of the policy parameters (typically the weights and bias of a neural network). A general algorithm for constrained policy optimization under a fixed set of backoff values is provided by Algorithm \ref{Alg:SafePO}. These backoffs are adjusted via BO - details are presented later in the manuscript in Section \ref{sec:BO} and Algorithm \ref{Alg:BOBackoffs}. 

% insert algorithm
\vspace{0.2cm}
\begin{algorithm}[h!]
\SetAlgoLined
\caption{Safe Policy Optimization for Fixed Backoffs}
\vspace{0.1cm}
\justify
\textbf{Initialise}: Experimental dataset $\mathcal{D}$; GP state space model $f_{GPSS} = [f^{1}_{GP}(\bm{\upsilon}), \ldots, f^{n_x}_{GP}(\bm{\upsilon})]$ with hyperparameters $\hat{\Lambda} = [\hat{\bm{\lambda}}_1, \ldots, \hat{\bm{\lambda}}_{n_x}]$ trained on $\mathcal{D} $; Initial control policy $\pi(\mathbf{u}|\mathbf{x}; \theta_0)$; Policy optimization algorithm $f_{PO}(\cdot)$; backoff multipliers $\bm{\xi}$; Finite horizon length T; initial state distribution $p(\mathbf{x}_0)$; Memory $\mathcal{B}_{info}$ for information required for $f_{PO}(\cdot)$; $K$ episodes; tolerance criterion\vspace{0.1cm}\\
\textbf{1.} $i = 0$\\
\textbf{2.}\ \While{not converged}{\vspace{0.1cm}
                    \textbf{a.} Obtain a batch of $K$ rollouts over a horizon of $T$ discrete intervals according to 
    
                    Algorithm \ref{Alg:UncertProp}, via $\pi(\mathbf{u}|\mathbf{x}; \theta_i)$, $f_{GPSS}$, and $p(\mathbf{x}_0)$. Return the trajectory information\footnotemark[3] of 
                    
                    each rollout and any further necessary information for $f_{PO}(\cdot)$ and store in $\mathcal{B}_{info}$.\vspace{0.1cm}\;
                    
                    \textbf{b.} Perform policy optimization $\theta_{i+1} = f_{PO}(\mathcal{B}_{info}, \theta_i)$\vspace{0.1cm}\;
                    
                    \textbf{c.} Reset memory  $\mathcal{B}_{info}$\vspace{0.1cm}\;
                    
                    \textbf{d.} i += 1\vspace{0.1cm}\;
                    
                    \textbf{e.} Assess tolerance criterion\vspace{0.1cm}\;}
\justify \textbf{3.} Assess final policy performance $J(\theta_i)$ under the unconstrained reward function $R$ and approximate the probability of joint constraint violation (Eq. \ref{eq:jointChance}) denoted $F_{LB}(0)$ via the method detailed in \ref{sec:jointSA}
\justify\textbf{Output:}  Optimal constrained policy $\pi^*_C(\mathbf{u}|\mathbf{x}; \theta_i)$ under backoff multipliers $\bm{\xi}$ and associated performance indices $J(\theta_i)$ and $F_{LB}(0)$ \vspace{0.1cm}
\label{Alg:SafePO}
\end{algorithm} 

The description provided by Algorithm \ref{Alg:SafePO}  considers all \textit{on-policy} policy optimization approaches, denoted generally as $f_{PO}(\cdot)$, although there is no reason the approach could not utilise an off-policy method too \cite{pan2020constrained}. The detail provided formalises the process of obtaining function space realisations from the GP state space model, $f_{GPSS}$, each of which represents a potential instance of the uncertain process detailed by Eq. \ref{eq:nonlinDynA}. Every process trajectory is ranked according to Eq. \ref{eq:PF} and the current iterate of backoff multiplier, $\bm{\xi}$, values. Using the collected experience (including relevant information that describes decision making), stored in the memory, $\mathcal{B}_{info}$, the weights of the policy are updated by $f_{PO}(\cdot)$. This is repeated until a convergence criterion is satisfied. In the following computational experiments detailed by this work, \textit{the methodology was integrated with the proximal policy optimization (PPO) algorithm} \cite{schulman2017proximal}. This is an attractive option given: a) the ability to directly parameterise a policy as a conditional probability distribution over a \textit{continuous} control input space; b) compatibility with recurrent neural networks \cite{schulman2017proximal}; c) sample efficiency relative to conventional policy optimization methods i.e. \textit{reinforce}; and, d) ease of implementation. Full detail of the PPO algorithm is provided by \ref{sec:PPOdetails}. 

In this section, the methodology has provided mechanism to incorporate information about the constrained problem into the reward signal characteristic of the MDP. In doing so, a strong connection to reward shaping is established. In reward shaping, policy invariant modifications of the reward function are identified to aid learning of the optimal policy $\pi^*$ \cite{ng1999policy, dong2020principled}. In this work, we construct a policy varying modification of the reward function in order to satisfy operational constraints. The resultant penalty function (Eq. \ref{eq:PF}) contains a number of free parameters.  In the subsequent section, it is proposed to tune the backoffs, $\bm{\varepsilon}$, via the multipliers, $\bm{\xi}$, (see Eq. \ref{eq:backoffCC}) and a BO scheme. This leaves decision as to the parameters $\kappa\in \mathbb{R}$ and $\zeta\in \mathbb{R}$ open to the implementation, although it is recommended that they are large real values \cite{nocedal2006numerical}. Ultimately this provides mechanism for the implementation to balance operational risk and performance manually. 
\footnotetext[3]{This includes the rewards $\mathbf{\varphi}^{(k)}_{0:T-1} = [\mathbf{\varphi}^{(k)}_1, \ldots, \mathbf{\varphi}^{(k)}_{T-1}]$ under Eq. \ref{eq:PF} and the current backoff multipliers $\bm{\xi}$, for the sequence of controls $\mathbf{u}^{(k)}_{0:T-1} = [\mathbf{u}^{(k)}_1, \ldots, \mathbf{u}^{(k)}_{T-1}]$ and states $\mathbf{x}^{(k)}_{0:T} = [\mathbf{x}^{(k)}_1, \ldots, \mathbf{x}^{(k)}_{T}]$}

\subsubsection{Optimization of Backoff Multipliers}\label{sec:BO}

The primary objective of this section is to identify a mechanism which facilitates synthesis of a policy that:
\begin{enumerate}[(a)]
        \item achieves high probability constraint satisfaction as desired, and
        \item performs with respect to the original process objective as specified by $R: \mathbb{X} \times \mathbb{U} \times \mathbb{X} \rightarrow \mathbb{R}$.
\end{enumerate}
'Efficient' global optimization of the backoff multipliers $\bm{\xi}$ is proposed, and so the methodology explores definition of an objective to evaluate candidate values of $\bm{\xi}$ as follows. 

Firstly, discussion is directed in how best to evaluate a) from the policy generated by Algorithm \ref{Alg:SafePO}. Specifically, with reference to Eq. \ref{eq:Boole} and the following works \cite{paulson2020stochastic, petsagkourakis2020chance}:
\begin{equation}
    \begin{aligned}
        F_X(0) &= \mathbb{P}(X \leq 0) = \mathbb{P}(\bigcap_{i=0}^T \{\textbf{x}_i \in \mathbb{\hat{X}}_i\})
    \end{aligned}\label{eq:jointprobcdf}
\end{equation}
where $F_X(\cdot)$ indicates the cumulative distribution function (cdf), which in this case is analytically intractable. In order to assess a), it is proposed to validate the probability of constraint satisfaction empirically via MC sampling under the GP state space model i.e. via the sample approximation of $F_X(0)$, denoted $F_{SA}(0)$. The specific approach is detailed in \cite{petsagkourakis2020chance} and repeated in \ref{sec:jointSA} for completeness. Ultimately, through this sampling-based method, a lower bound for Eq. \ref{eq:jointprobcdf} and $F_{SA}(0)$ is obtained and denoted $F_{LB}(0)$. This accounts for potential inexactness introduced through finite samples. 
The evaluation of b) is more simple and directed via the definition of the process objective in the form of the reward function $R$. As such, the investigation may evaluate the performance of the policy under the original, unconstrained objective provided by Eq. \ref{eq:objective}. This work therefore proposes the use of the following objective function in evaluation of candidate multiplier values $\bm{\xi}\in \mathbb{R}^{n_g}$:
\begin{equation}
    \begin{aligned}
        \bm{U} &= (F_{LB}(0) - (1-\alpha))^2 \\
        J_{BO} &= -(J(\bm{\tau}) - \beta \sigma_{J})\exp(-c\bm{U})
    \end{aligned}
    \label{eq:JBO}
\end{equation}
where $\beta = [0,1]$, $c \in \mathbb{R}^+$ and $\sigma_{J}$ denotes the standard deviation of the policy with respect to the unconstrained process objective. This is a modification to the objective function proposed previously in \cite{petsagkourakis2020chance}, which equated $J_{BO}$ to $\bm{U}$. Here, Eq. \ref{eq:JBO}, provides a smoother latent function and naturally balances the objectives a) and b). The inclusion of the term $\sigma_{J}$ also incentivises those policies, which exploit regions of the state space well characterised by the model. The factor $c$ provides a shape parameter for the RBF part of the objective, with higher values providing greater incentive to obtain joint constraint satisfaction as desired. Care should be taken in selection as the higher the value, the sparser the mapping provided by the objective. This is likely to have consequences for the efficacy of optimization. 

%This conclusion was drawn via preliminary experiments and modelling of the composite function $\bm{U}$, which is explicitly dependent upon the backoff multiplier values, $\bm{\xi}$.%
%
%begin{figure}[h!]
  %  \centering
   % \subfigure(a){%
 %       \includegraphics[width=7cm,height=5cm,keepaspectratio]{Figures/BO/rbffunction.png}
  %      }
   % \subfigure(b){%
    %    \includegraphics[width=7cm,height=5.cm,keepaspectratio]{Figures/BO/model_landscape.png}
     %   }
    %\caption{Visual demonstration of the concepts discussed in Section \ref{sec:BO}. a) The effect of tuning the shape parameter $c$ on the mapping provided by the radial basis function compositional to Eq. \ref{eq:JBO}. b) Visualisation of the optimisation landscape from a \textit{toy} backoff multiplier ($\bm{\xi}$) optimisation problem with one constraint ($n_g =1$). The problem is underpinned by modelling assumptions regarding the influence of the backoff multiplier $\bm{\xi}$ on the probability of constraint satisfaction, $F_{LB}(0)$. These assumptions were developed empirically from preliminary experiments on the real problem. In this case, the objective, $J_{BO}$, was defined with $c=1$. It can be seen that the objective provides a clear preference over backoff multipliers and a smooth latent function. This plot was generated via the GPyOpt package.}
    %\label{fig:BOassumptions}
%\end{figure}
In the following case studies, $\beta = 0.1, c = 1$ and a BO scheme was deployed via GP surrogate models with RBF covariance functions and zero mean priors to optimize the backoff multipliers $\bm{\xi}$:
\begin{equation}
\begin{aligned}
    \bm{\xi}^* = \argmin_{\bm{\xi}} J_{BO}
\end{aligned}
\end{equation}
Due to the expensive black box optimization proposed BO is deemed the most appropriate approach. BO proceeds to construct and exploit a GP surrogate model to sample new candidate points. Construction of the GP surrogate demands a small initial dataset, describing a set of inputs, $\bm{\Xi}$, and their corresponding fulfilment of the objective function, $J_{\bm{\Xi}}$. New sampling points (or in this case, candidate backoff multipliers) are sampled to maximise an acquisition function (AF), which is a function of the posterior distribution of the GP surrogate. The AF, denoted $f_{AF}(\cdot)$, used in this work was the expected improvement (EI) function  \cite{frazier2018tutorial,jones1998efficient}. It was found that the EI AF, $f_{AF}^{EI}(\cdot)$, was most efficient in this case, balancing exploration and exploitation of the GP surrogate model to find the optimal solution. The expected improvement function is detailed as follows \cite{jones1998efficient, brochu2010tutorial}:
\begin{align*}
    \varrho &= \frac{\mu(\bm{\xi}) - J_{BO}^{+}}{\sigma(\bm{\xi})}
\end{align*}
\begin{equation}\label{eq:EI}
    \begin{aligned}
        f_{AF}^{EI}(\bm{\xi}) &= \begin{cases}
    \sigma(\bm{\xi}) \psi(\varrho ) + (\mu(\bm{\xi}) - J_{BO}^{+})\phi(\varrho),& \text{if } \sigma(\bm{\xi}) > 0\\
    0,              & \text{otherwise}
\end{cases} \\
    \end{aligned}
\end{equation}
where $\phi(\cdot)$ is the Gaussian cumulative distribution function, $\psi(\cdot)$ is the Gaussian probability density function, $J_{BO}^{+}$ is the objective value of the current best backoff multiplier values $\bm{\xi}^+$ \cite{jones1998efficient, frazier2018tutorial}, and $\mu(\cdot)$ and $\sigma(\cdot)$ are detailed by Eq. \ref{eq:posterior}. Dissecting Eq. \ref{eq:EI}, the first term on the right hand side incentivises exploring regions of the input space associated with high uncertainty in the posterior distribution, and the second term provides basis to exploit regions of the input space corresponding to high mean predictions in the posterior \cite{jones1998efficient}. As such, Eq. \ref{eq:EI} provides explicit mechanism to balance exploration and exploitation of the GP surrogate model, in a fashion not dissimilar to the exploration-exploitation paradigm in RL. 
For more detail on BO in this context, we direct the interested reader to previous work \cite{petsagkourakis2020chance, del2021real} and a comprehensive review \cite{frazier2018tutorial}.

Algorithm \ref{Alg:BOBackoffs} formalises the approach to reward shaping detailed by this Section. In \textbf{Step 1}, an optimal policy parameterisation is learned for the unconstrained problem. This is used as an initialisation for learning of the optimal constrained policy thereafter. In \textbf{Step 2.a}, a number of policies are learned for the constrained problem each utilising different values of the backoff multipliers. In \textbf{2.b}, each of the policies is assessed with respect to $J_{BO}$, providing an input-output dataset, where the inputs are backoff multipliers and the outputs are corresponding performances under the objective  ($J_{BO}$). In \textbf{Step 3}, a surrogate GP model is built via this input-output dataset for subsequent BO. Pseudocode for BO is provided by \textbf{Step 4}, with \textbf{Step 5} documents the return of the solution policy from memory. 

\vspace{0.2cm}
\begin{algorithm}[H]
%\SetAlgoLined
\caption{Safe Chance Constrained Policy Optimization}
\vspace{0.1cm}
\justify 
\textbf{Initialisation}: Desired probability of joint chance constraint satisfaction $\alpha$; GP prior for Bayesian Optimization $f_{BO}$; Acquisition function $f_{AF}$; Objective function $J_{BO}$; maximum number of acquisitions for BO $M$; Initial set of $B$ backoff multiplier values $\bm{\Xi} = [\bm{\xi}_1, \ldots, \bm{\xi}_B]$ generated via sobol sequence \cite{sobol1967distribution}; \vspace{0.1cm} \\
\textbf{1}. Perform policy optimization for unconstrained problem to maximise Eq. \ref{eq:objective} via modification to Algorithm 2. Return policy $\pi^*(\cdot,\theta)$.\vspace{0.1cm}\\
\textbf{2.a.} Train a set of $B$ constrained policies $\pi^*_{init} = [\pi^*_C(\cdot,\theta_1), \ldots, \pi^*_C(\cdot,\theta_B)]$ to maximise Eq. \ref{eq:consobjective} under the respective backoff values, $\bm{\Xi}$, via Algorithm 2 with $\pi^*(\cdot,\theta)$ for initialisation \vspace{0.1cm}\\
\textbf{2.b.} Return performance indices $F_{LB}(0)$ and $J(\bm{\tau}, \theta) \ \forall \ \pi^*_C(\cdot, \theta) \in \pi^{*}_{init}$ and assess $J_{BO}$, such that $J_{\Xi} = [J_{BO}(\bm{\xi}_1), \ldots, J_{BO}(\bm{\xi}_B)]$.\vspace{0.1cm} \\
\textbf{3.} Train a GP model given input-output pairs representative of (backoff multiplier values and policy performance under $J_{BO}$) $\bm{\Xi}\ \text{and} \ J_{\bm{\Xi}}$ according to \ref{sec:GPtrain} and condition to obtain updated predictive posterior distribution, $p(J_{BO}| \bm{\xi}, \bm{\Xi})$.\vspace{0.1cm}\\
\textbf{4.}\ \For{m = 1, \ldots, M}{\vspace{0.1cm}
                            \textbf{a.} According to $p(J_{BO}| \bm{\xi}, \bm{\Xi})$ find $\bm{\xi}_{B+m} = \argmax_{\bm{\xi}} f_{AF}(\cdot)$ and update 
                            
                            $\bm{\Xi} = [\bm{\xi}_{1}, \ldots, \bm{\xi}_{B+m}]$
                            
                            \textbf{b.} Train constrained policy $\pi^*_C(\cdot,\theta_{B+m})$ via Algorithm 2, $\pi^*(\cdot,\theta)$ for initialisation under 
                            
                            the backoff values $\bm{\xi}_{B+m}$. Return performance indices $F_{LB}(0)$ and $J(\bm{\tau}, \theta_{B+m})$ \text{for} 
                            
                            $ \pi^*_C(\cdot, \theta_{B+m})$, assess $J_{BO}(\bm{\xi}_{B+m})$ and append to dataset, $J_{\Xi} = [J_{BO}(\bm{\xi}_1), \ldots, J_{BO}(\bm{\xi}_{B+m})]$\vspace{0.1cm}
                            
                            \textbf{c.} \textit{if} $m<M$: repeat step \textbf{3.}
                            }
\justify\textbf{5.} Return $\pi_C^*(\theta)$ corresponding to $\bm{\xi}^* = \argmax_{\bm{\xi}}J_{\bm{\Xi}}$ \\
\textbf{Output:} Optimal Constrained Policy $\pi_C^*(\theta)$
\label{Alg:BOBackoffs}
\end{algorithm} 
\vspace{0.2cm}
In the following section, the method is demonstrated on a microalgal lutein photo-production dynamic process and benchmarked against dynamic optimization and NMPC strategies.

\section{Case Study}\label{sec:CS}
To demonstrate the methodology, a case study was selected from previous work conducted by \cite{zhang2019hybrid, del2017kinetic}, which is underpinned by a set of ordinary differential equations (ODEs). The problem and standard benchmarks are detailed via the following subsections.
\subsection{A Microalgal Lutein Photo‐Production Dynamic Process}\label{sec:luteinCS}
Fed-batch fermentation processes are thought to be ideal systems for RL-based controllers and particularly suited to data-driven approaches to control and optimization. This is due to characteristics of predominantly batch mode operation and complex physical phenomena driven by the metabolic reaction network. The complexity of the process physics often provides impediment to structural and practical model identification, with large parametric uncertainties common across bioprocess systems. To demonstrate the method proposed here, we consider an \textit{in-silico} microalgal lutein photo-production process described as follows: 
\begin{equation}\label{eq:LSYS}
    \begin{aligned}
        \dot{c}_X &= u_0\frac{c_N}{c_N + K_N} c_X - u_d c_X \\
        \dot{c}_N &= - Y_{N/X} u_0\frac{c_N}{c_N + K_N} c_X + F_{N,_{in}} \\
        \dot{c}_L &= k_0\frac{c_N}{c_N + K_{NL}} c_X - k_d c_Lc_X \\
    \end{aligned}
\end{equation}
where $c_X \ (g \ L^{-1})$ defines the biomass concentration; $c_N \ (mg \ L^{-1})$ defines the nitrate concentration; $c_L\ (mg \ L^{-1})$ defines the lutein (product) concentration; $F_{N,_{in}} (mg \ h^{-1})$ is the nitrate inflow to the system (a control input); $u_0 \in \mathbb{R}$ ($h^{-1}$) is the specific biomass growth rate, which is a function of the incident light intensity $I_0\in \mathbb{R}$ ($\mu mol \ m^{-2} \ s^{-1} $) to the reactor and the maximum theoretical growth rate $u_m \in \mathbb{R}$ ($h^{-1}$); $k_0 \in \mathbb{R} $ ($mg \ g^{-1}\ h^{-1}$) is the specific lutein production rate, which is a function of $I_0$  and the maximum theoretical production rate $k_m \in \mathbb{R}$ ($mg \ g^{-1}\ h^{-1}$); $k_d \in \mathbb{R}$ ($L \ g^{-1} \ h^{-1}$) is the lutein consumption rate; $u_d \in \mathbb{R}$ ($h^{-1}$) is the biomass specific decay rate; $Y_{N/X} \in \mathbb{R}$ ($mg \ g^{-1} $) is the nitrate yield coefficient; and, $K_N \in \mathbb{R}$ ($mg \ L^{-1}$) and $K_{NL} \in \mathbb{R}$ ($mg \ L^{-1}$) are the nitrate half-velocity constant for cell growth and lutein synthesis, respectively. The growth rates of biomass and lutein are constituted by the terms  $u_0$ and $k_0$. These are both functions of the incident light intensity to the reactor and are detailed as follows:
\begin{equation}\label{eq:kinetics}
    \begin{aligned}
        u_0  &= \frac{u_m}{20} \sum_{n=1}^9\bigg( \frac{I_0}{I_0 +k_s +\frac{I_0^2}{k_i}} + 2\frac{I_\frac{nL}{10}}{I_\frac{nL}{10} +k_s +\frac{I_\frac{nL}{10}^2}{k_i}} + \frac{I_L}{I_L +k_s +\frac{I_L^2}{k_i}}\bigg) \\
        k_0  &= \frac{k_m}{20} \sum_{n=1}^9\bigg( \frac{I_0}{I_0 +k_{sL} +\frac{I_0^2}{k_{iL}}} + 2\frac{I_\frac{nL}{10}}{I_\frac{nL}{10} +k_{sL} +\frac{I_\frac{nL}{10}^2}{k_{iL}}} + \frac{I_L}{I_L +k_{sL}} +\frac{I_L^2}{k_{iL}}\bigg)
    \end{aligned}
\end{equation}
where $k_i \in \mathbb{R}$ and $k_{iL}\in \mathbb{R}$ are light inhibition terms for biomass growth and lutein synthesis, respectively. Similarly, $k_s\in \mathbb{R}$ and $k_{sL}\in \mathbb{R}$ are light saturation terms for biomass growth and lutein synthesis, respectively. More information on parameter definitions and values is provided by \cite{del2017kinetic, zhang2019hybrid}. The states $\mathbf{x} = [c_X, c_N, c_L ]$ of the system are absolute and hence $c_i \geq 0 \ \forall \ i\in\{X, N, L\}$. Further to $F_{N,_{in}}$, the control input is also constituted by $I_0$, such that $n_u = 2$ and $\mathbf{u}(t) = [F_{N,_{in}}, I_0] ^T$, with bounds $ 0.1 \leq F_{N,_{in}} \leq 100 $ $mg \ h^{-1}$  and $100 \leq I_0 \leq 1000$ $\mu mol \ m^{-2} \ s^{-1} $. It is assumed that the process is subject to stochasticity in the form of 5 \% parametric uncertainty. This and the initial state distribution is detailed by Table \ref{table:uncertainBP}. Parametric values not detailed are assumed constant following the original works \cite{del2017kinetic, zhang2019hybrid}. Additionally, the initial state distribution, $p(\mathbf{x}_0)$, is defined in keeping with the work \cite{zhang2019hybrid}.
\begin{table}[h!]
  \caption{Case Study: List of parametric and initial state distributions imposed to describe uncertainty in the real underlying bioprocess.}
  \label{table:uncertainBP}
  \centering
  \begin{tabular}{llll}
    \toprule

    Variable & Uncertainty Distribution \\
    \midrule
    $u_m$ & $\mathcal{N}(0.152, 0.0038)$  \\
    $K_N$ & $\mathcal{N}(30, 0.75)$  \\
    $u_d$ & $\mathcal{N}(5.93 \times 10^{-3}, 1.483 \times 10^{-4})$\\
    $Y_{N/X}$ & $\mathcal{N}(305, 7.625)$ \\
    $k_m$ & $\mathcal{N}(0.35, 0.00875)$\\
    $K_d$ & $\mathcal{N}(3.71\times10^{-3}, 9.275 \times 10^{-5})$\\
    $\mathbf{x}_0$ & $\big[\mathcal{N}(0.27, 3.125\times 10^{-3} ), \mathcal{N}(765.0, 9.5625), \mathcal{N}(0.0,0.0)\big]$\\
    \bottomrule
  \end{tabular}
\end{table}
However, the state constraints imposed are specific to this work and not the previous. Here, affine constraints are defined using notation from Section \ref{sec:constrainedPPO}:
\begin{equation}\label{eq:cons}
    A = \begin{bmatrix} 1 & 0 & -1.67 \\
                       0 & -1\times 10^{-3} & 0 \\
                       0 & 0 & 1\end{bmatrix} \qquad b = \begin{bmatrix} 2.6 \\ 0.15 \\ 0 \end{bmatrix}
\end{equation}
The constraints were constructed to represent common operational concerns in bioprocessing. The first column of $A$ considers the potential raw material to product conversion via constraint of the maximum biomass concentration (as biomass is a 'by-product'). The second column considers the protection of cell growth (via a minimum nitrate constraint) and the third ensures continued productivity (via constraint of the maximum ratio of secondary metabolite to biomass).  The process objective reward function $R: \mathbb{X} \times \mathbb{U} \times \mathbb{X} \rightarrow R_{t+1}$ is as follows:
\begin{equation}
    \begin{aligned}
        R_{t+1} &= \begin{cases} \mathbf{d}^T \mathbf{x}_{t+1} - 
    \Delta\mathbf{u}_t^TC\Delta\mathbf{u}_t & \text{if } t = T-1\\
    - \Delta\mathbf{u}_t^TC\Delta\mathbf{u}_t,              & \text{otherwise}
\end{cases} \\
    \end{aligned}\label{eq:CSObj}
\end{equation}
where $t = [0, \ldots, T]$ and the length of the finite horizon is defined $T = 6$; $\Delta\mathbf{u}_t = \mathbf{u}_t - \mathbf{u}_{t-1} \in \mathbb{R}^{n_u} $ defines the change of controls between discrete time steps;  $C = diag([0.16, 8.1\times 10^{-5}]) \in \mathbb{R}^{n_u \times n_u}$ provides a penalty for changing the controls and promotes the learning of 'stable' control profiles; and, $d = [0, -0.001, 4]^T \in \mathbb{R}^{n_x}$ provides an overall objective for process operation i.e. to maximise the production of lutein and minimise waste of nitrate.  The problem definition is common to both Section \ref{sec:CSsafeCCPO} and the benchmark described in Section \ref{sec:Bench}, except the benchmark does not consider any form of parametric uncertainty. A formalisation of the control problem follows:
\begin{equation}
\mathcal{P}(\pi_C):=\left
\{\begin{aligned}
        &\max_{\pi_C} \mathbb{E}_{\pi_C}\bigg[\sum_{t=0}^{T-1} R_{t+1}\bigg] \quad \quad \quad \quad  (\text{see Eq. \ref{eq:CSObj}})\\
    &\text{s.t.}\\
    &\textbf{x}_0 \sim p(\textbf{x}_0)\\
    & \textbf{s}_t \sim p(\textbf{s}) \quad \quad \quad \quad \quad \quad \quad \quad  \ \quad (\text{see Table \ref{table:uncertainBP}} )\\
    &\textbf{x}_{t+1} = f(\textbf{x}_t, \textbf{u}_t, \textbf{s}_t) \quad \quad \quad \quad \quad (\text{see Eqs. \ref{eq:LSYS} and \ref{eq:kinetics}})\\
    &\textbf{u}_t = \pi_C(\textbf{x}_t)\\
    &\textbf{u}_t\in\mathbb{\hat{U}}\\
    &\textbf{x}_t \in \mathbb{\hat{X}}_t \quad \quad \quad \quad \quad \quad \quad \quad \quad \quad (\text{see Eqs. \ref{eq:safeset} and \ref{eq:cons}})\\
    &  \forall t \in \left\{0,...,T-1\right\}\label{eq:CSOCP}
\end{aligned}\right.
\end{equation}
\subsection{Safe Chance Constrained Policy Optimization}\label{sec:CSsafeCCPO}
To demonstrate the methodology, this work deploys the PPO algorithm with both actor and critic recurrent long-short term memory (LSTM) neural network parameterisations. The actor network expresses a mapping between observed states and controls (i.e. a control policy) and the critic provides a mapping between a state and the value of that state under the policy (this is known as the value function). The use of a critic provides means to deploy the general advantage estimate (GAE) form of the policy gradient (PG) within the PPO framework. The GAE enables the implementation to manually balance the bias and variance of the advantage PG. This provides means to synchronously ensure stable learning, improve the sample efficiency of the algorithm and find a clear direction (in weight space) for policy improvement. For more information on PPO and the GAE, the reader is directed to \ref{sec:PPOdetails} and \cite{schulman2017trust, schulman2017proximal}. The implementation utilised Pytorch 1.7.1. Information about the structure of the actor, critic and all hyperparameters defining the PPO algorithm as used in this work, may be found in \ref{sec:EntropyRegPPO}. See Table \ref{table:uncertainBPHP} for definition of general case study parameters.
\begin{table}[h!]
  \caption{Case Study: List of key algorithm parameters}
  \label{table:uncertainBPHP}
  \centering
  \begin{tabular}{llll}
    \toprule
    Variable & Value \\
    \midrule
    Penalty weight, $\kappa$ & $34$  \\
    Uncertainty penalty weight, $\zeta$ & $300 \times diag([1/\sigma^2_{1_{\Upsilon}}, \ldots, 1/\sigma^2_{{n_x}_\Upsilon}])$\footnotemark[4] \\
    Tolerance criterion & $|\bar{J}_C(\bm{\tau}, \theta_i) - \bar{J}_C(\bm{\tau}, \theta_{i-1})| \leq 10^{-3}$\\
    Joint Probability of constraint violation, $\alpha$ & $0.001$\\
    Probability of individual constraint violation, $\iota_j$ & $0.00033$\\
    \bottomrule
  \end{tabular}
\end{table}
\footnotetext[4]{ $\sigma^2_{{n_x}_\Upsilon}$ represents the variance of the distribution of state $x_{n_x}$ in the dataset $\mathcal{D}$} 
\footnotetext[5]{ The dataset used for model construction may be found at https://github.com/mawbray/Lutein-Dataset}

In order to train the desired policy $\pi_C^*(\cdot, \theta)$ via Algorithm \ref{Alg:BOBackoffs}, a GP state space model is required. In this work, the model was built using an initial dataset $\mathcal{D}$\footnotemark[5], generated by simulation of the uncertain process' response to 32 different control sequences, $\mathbf{u}_{0:T}^{(j)}, \ j = [1, \ldots, 32]$ (hence the dataset contains information from 32 separate batch experiments). Each control sequence was generated via transformation of a Sobol sequence (of length T) to the bounded controls space (as detailed in Section \ref{sec:luteinCS}).  In practice, this dataset could be generated via an initial design of experiments \cite{petsagkourakis2020safe}. Having generated $\mathcal{D}$, ($n_x = 3$) individual GP models were constructed to form a state space model (for the prediction of each state) via the methodology outlined in Section \ref{sec:GPdyna}. A prior distribution with mean function $\mathbf{m}(\bm{\upsilon}) = 0$ and a matern 5/2 covariance function was specified for each of the constituent models. The covariance function was selected according to preliminary experiments, which examined the model's predictive accuracy. All GPs were constructed with the GPy 1.9.9 python package and subsequent BO utilised GPyOpt 1.2.6. Details of the data used for model construction, as well as metrics relating to the predictive accuracy of the model are detailed in \ref{sec:GPvalid}.

In the presentation of results for this work, the investigation is concerned with two main questions. Firstly, does Algorithm \ref{Alg:BOBackoffs} enable identification of a reward function, which provides policy performance with respect to the process objective and probabilistic constraint satisfaction? And, secondly, does the incorporation of the posterior variance prediction of the GP state space model (into the reward function (Eq. \ref{eq:PF})) provide means to minimise the risk of policy deployment (to the real uncertain process), by ensuring the policy exploits regions of the model with small model-process mismatch? These two questions will direct discussion in Section \ref{sec:R&D}. All results were generated under view of the policy as deterministic i.e. $\mathbf{u}_t = \pi(\mathbf{x}_t)$. This was achieved through selection of the control corresponding to the mode of the conditional distribution $\pi(\mathbf{u}|\mathbf{x})$. 

\subsection{Benchmark for Process Optimization}\label{sec:Bench}
The results from the proposed methodology were benchmarked relative to the control profiles generated from a) dynamic optimization (DO) strategies, and b) nonlinear model predictive control (NMPC). Both a) and b) use the process model detailed by Eq. \ref{eq:LSYS}. The deterministic form of this model (i.e. with no parametric uncertainty) represents the most accurate deterministic model, which may be built for process prediction and optimization. Therefore, the controls generated from a) and b) assume that the underlying process is deterministic, and are subsequently validated on the stochastic analogue of the process concerned.  As both a) and b) neglect the existence of uncertainty over the parameter values assumed from \cite{del2017kinetic}, validation of the strategies on the stochastic variant of the process (detailed by Section \ref{sec:luteinCS}) directly investigates the effects of uncertainty (model-plant mismatch) on performance with respect to the objective and constraint satisfaction. It should be noted that this benchmark is not reflective of the existing state-of-the-art optimization methods, such as sMPC that similarly consider model uncertainty. The control strategy for a) was generated offline through optimization of the control inputs to the model detailed in Eq. \ref{eq:LSYS}. Hence the control policy generated is deterministic and unconditional to online state observation. Conversely. the control strategy for b) was generated online through perfect state observation as in the RL case. Both benchmarks utilised the orthogonal collocation method and one finite element per control interval \cite{biegler2007overview, kelly2017introduction} and the IPOPT solver \cite{wachter2006implementation}. This was facilitated by the Casadi 3.5.1 Python package \cite{andersson2012casadi}. In the case that a feasible solution could not be found online, the MPC scheme was tuned further with an approximate problem solved to minimise constraint violation. From empirical analysis, this tuning increased the performance of the NMPC scheme. Further information on the approximate problem is available in \ref{sec:benchmod}.

\subsection{Key Performance Indicators}
In the following section, this work will investigate the utility of the algorithm, and presentation of the results will focus on the ability of the proposed method to find a safe constrained policy $\pi_C^*(\cdot, \theta)$. Explicitly, the policy should exploit regions of the real process state space (i.e. Eq. \ref{eq:LSYS}), well characterised by the approximating process model (i.e. Eq. \ref{eq:GPSS}), therefore minimising mismatch between the state distributions simulated under the offline process model and observed under the real uncertain process. This will be demonstrated in two ways. First, via visual comparison as presented figuratively, and secondly via the quantitative metrics (key performance indicators) available to the investigation. Primarily, these metrics are the performance of the policy with respect to the unconstrained process objective $J(\tau)$ (see Eqs. \ref{eq:objective} and \ref{eq:CSObj}) and the probability of joint chance satisfaction as evaluated by $F_{SA}(0) \ \text{and} \ F_{LB}(0)$. The same metrics will be used to evaluate the performance of the benchmarks of DO and NMPC.

\section{Results and Discussion}\label{sec:R&D}
\subsection{Results of Safe Chance Constrained Policy Optimization}
\begin{figure}[h!]
    \centering
    \subfigure(a){\includegraphics[scale=0.223]{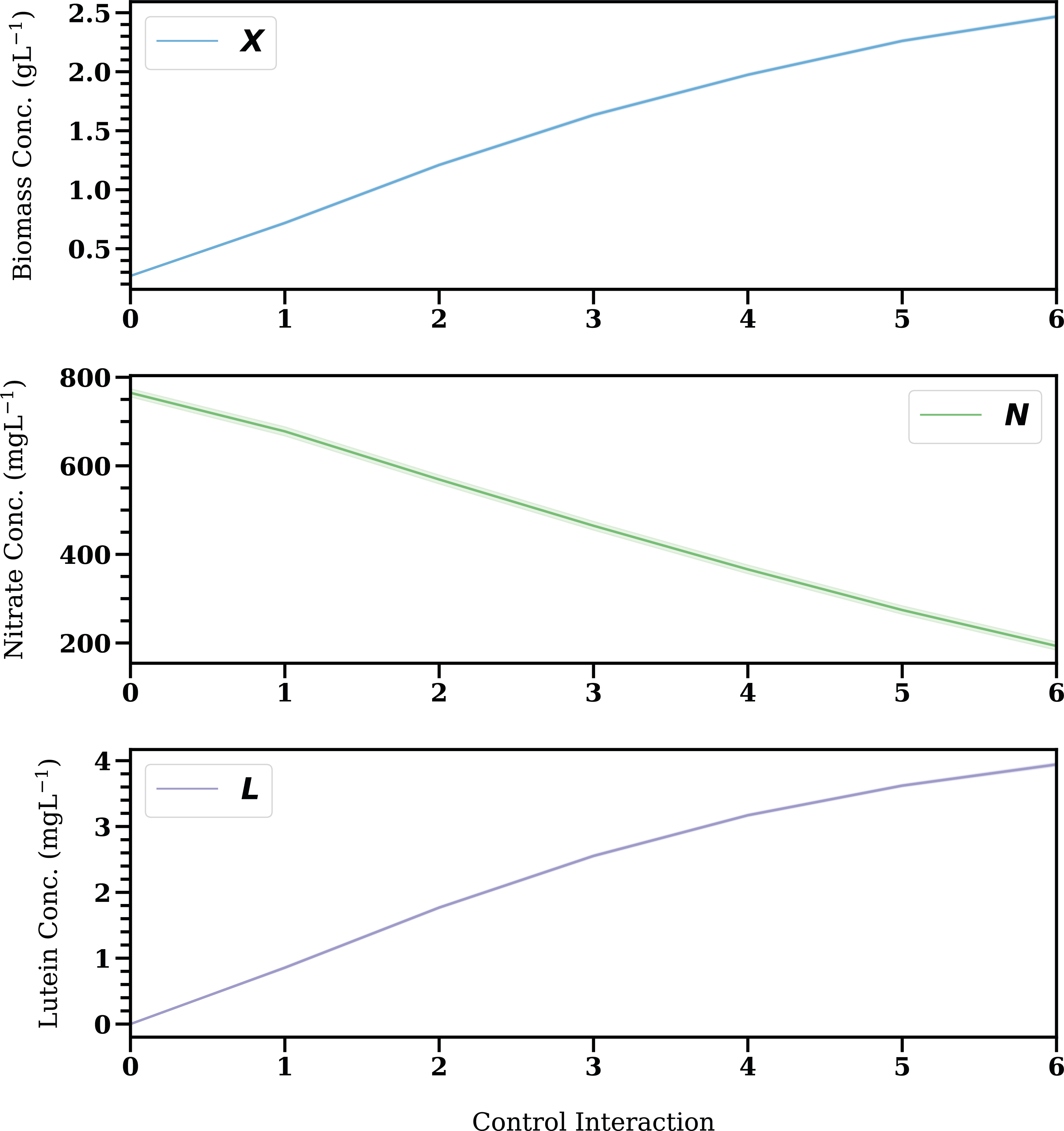}
        }
    \subfigure(b){
        \includegraphics[scale= 0.223]{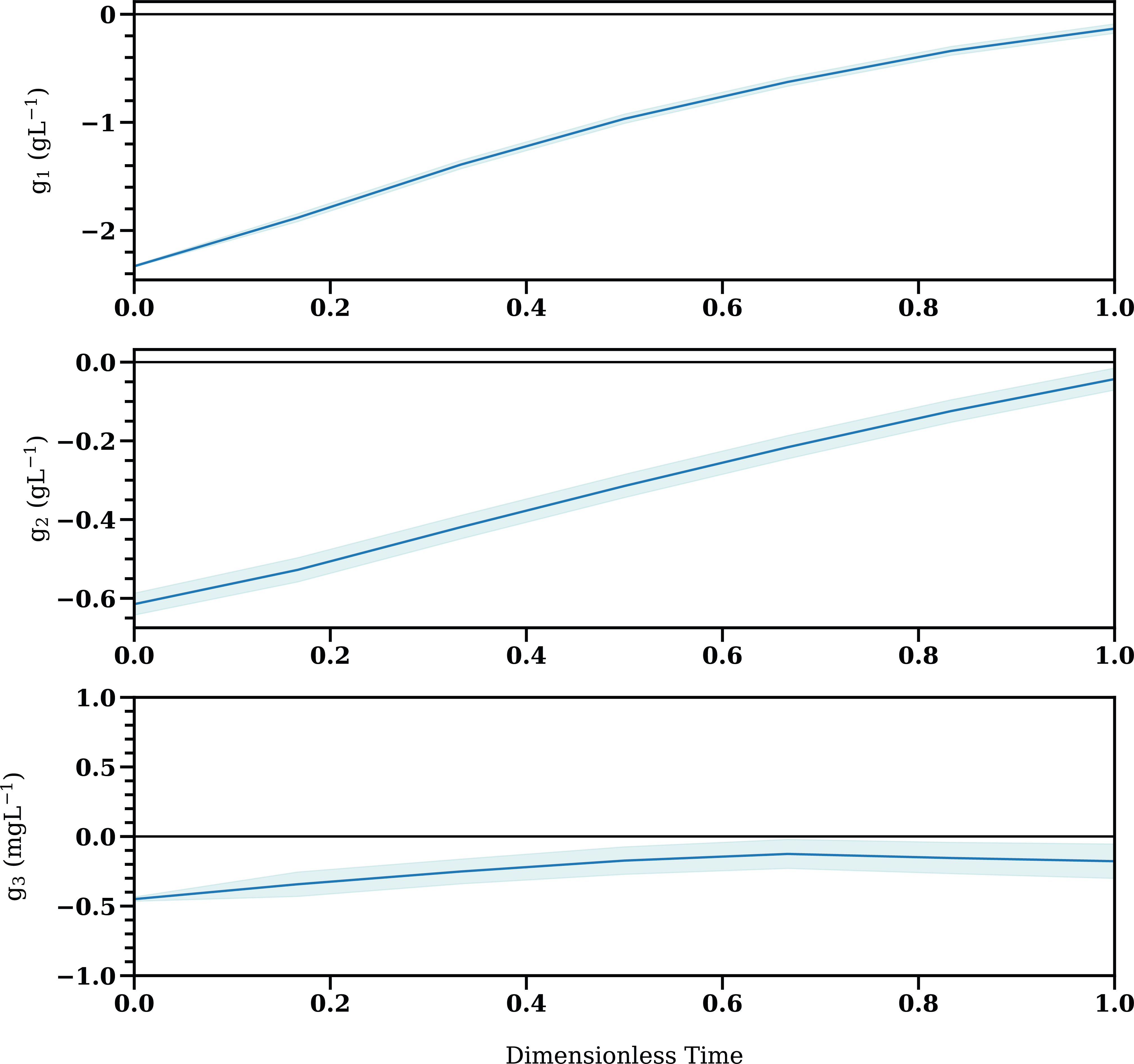}
        }
    \caption{Results from Case Study. (a) The state profile produced from the final policy learned on the Gaussian Process model plotted against control interactions (as a proxy for time). Control interactions are provided every 24 hours of process operation. (b) The corresponding distribution of trajectories with respect to the operational constraints. The $i^{th}$ constraint is denoted $g_i := A_i^T\mathbf{x} - b_i$. The light blue shaded areas represent the 99th to 1st percentiles and solid blue line represents the expected trajectory. The black line plot represents the threshold of constraint violation i.e. when $g_i = 0$}
    \label{fig:GPSSpolicy}
\end{figure}

\begin{figure}[h!]
    \centering
        \subfigure(a){%
            \includegraphics[scale=0.21]{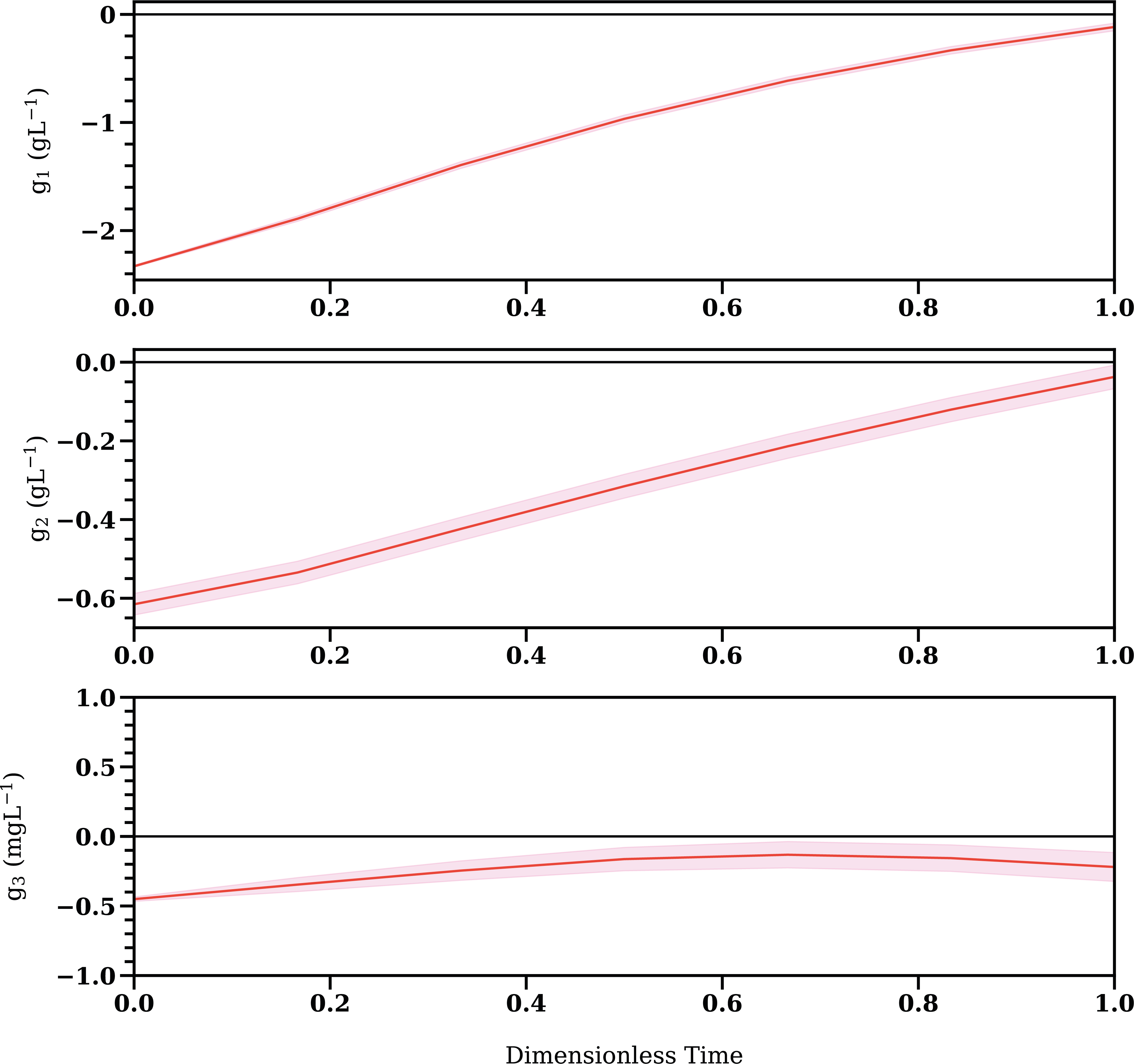}
            }
        \subfigure(b){%
            \includegraphics[scale= 0.21]{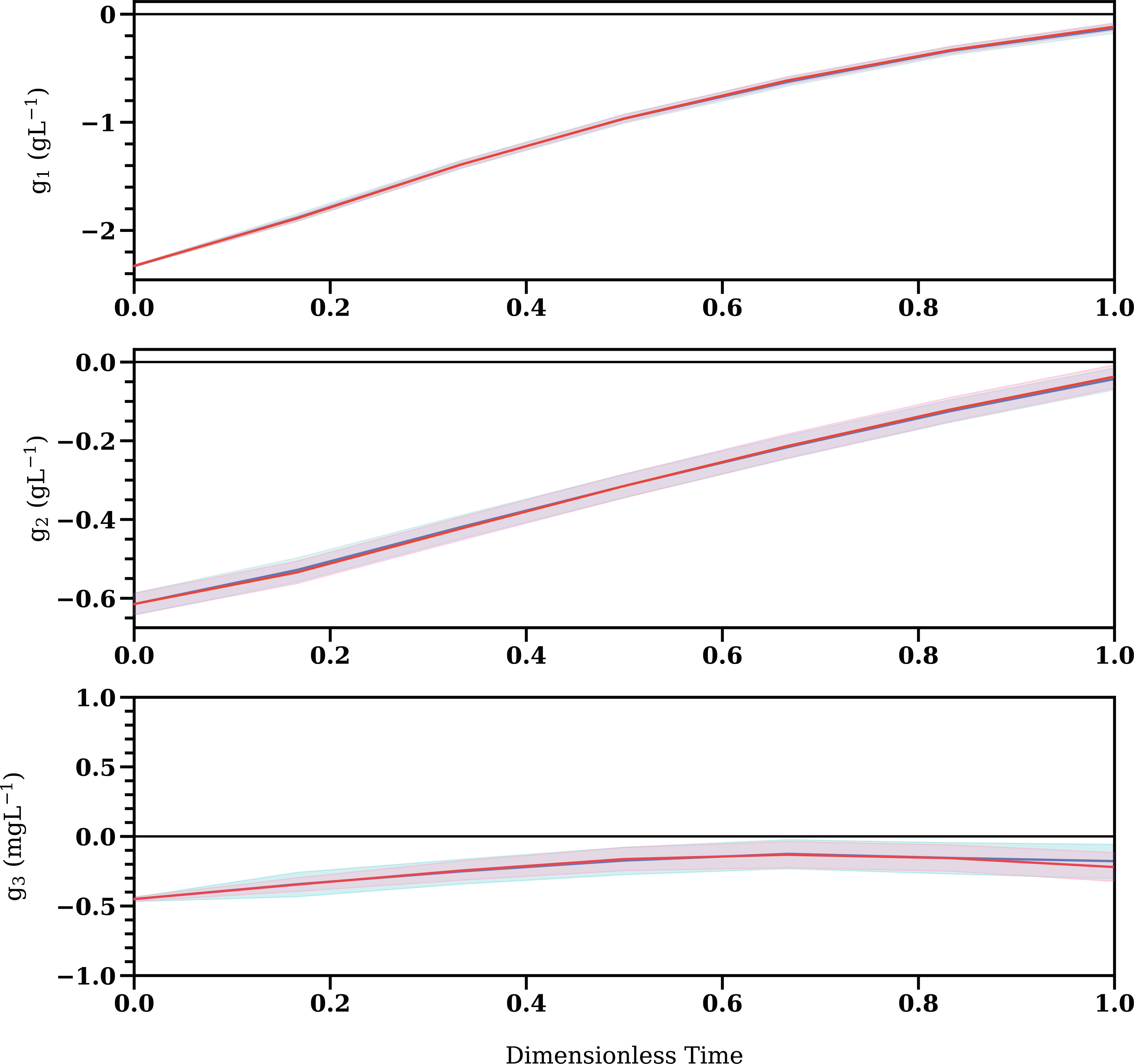}
            }
        \caption{Results from Case Study. (a) The distribution of trajectories with respect to the operational constraints as sampled from the real  uncertain process. (b) An overlay of the distributions observed when the policy is deployed on the real uncertain process (red) and the GP state space model (blue) as plotted in Fig. \ref{fig:GPSSpolicy}. The $i^{th}$ constraint is denoted $g_i := A_i^T\mathbf{x} - b_i$. The shaded areas represent the 99th to 1st percentiles and solid line represents the expected trajectory. The black line plot represents the threshold of constraint violation i.e. when $g_i = 0$}
        \label{fig:GPSS_real}
\end{figure}

Firstly, the results of Algorithm \ref{Alg:BOBackoffs} with respect to the approximate offline state space model are displayed by Fig. \ref{fig:GPSSpolicy}. Explicitly, here, we demonstrate the performance of the final policy $\pi_C^*(\cdot, \theta)$ on the \textit{GP state space model}. The results were obtained according to 500 function realisations of $\pi_C^*(\cdot, \theta)$ via Algorithm \ref{Alg:UncertProp}. Fig. \ref{fig:GPSSpolicy} a) expresses a representation of the state evolution $\mathbf{x}_{0:T}$ and Fig. \ref{fig:GPSSpolicy} b) provides a visualisation of the performance of the policy with respect to the constraints. In Fig. \ref{fig:GPSSpolicy} a), the average state evolution and an associated confidence interval of one standard deviation for the validation trajectories is represented by a solid line and a shaded region, respectively. It can be seen that the agent learns to maximise the productivity objective - balancing maximisation of the lutein product at the end of the batch with a decrease in the concentration of nitrate left in the system. This is achieved in a manner that accounts for worst case process stochasticity by backing the nominal or expected state trajectory away from the constraint boundary. This is highlighted by Fig. \ref{fig:GPSSpolicy} b). In particular, the shaded regions indicate 99\% confidence intervals for process deviation and the dark blue solid line plot indicates the nominal process. Further, the utility of tuning the backoff multipliers via Algorithm \ref{Alg:BOBackoffs} is highlighted given that the worst case realisations of process stochasticity do not violate, but approach the constraint boundary very closely.  The performance of the policy $\pi_C^*(\cdot, \theta)$ with respect to both process objective and constraint satisfaction on the GP process model is detailed by Table \ref{table:safeCCPO}. 
The performance of the policy on the process model is however, not the primary contribution of this work. Rather, it is of interest to validate the safety of the policy on deployment to the real stochastic process and highlight the particular use of the training approach detailed. To achieve this, results were obtained by sampling the real process described by Eq. \ref{eq:LSYS}, with the parametric uncertainty detailed in Section \ref{sec:CS}. The results of this are expressed by Fig. \ref{fig:GPSS_real}.
Similar to Fig. \ref{fig:GPSSpolicy} b), Fig. \ref{fig:GPSS_real} a) details the performance of the policy with respect to the constraints, but upon deployment to the real uncertain process. Again, the shaded regions indicate 99\% confidence intervals of process deviation and the dark blue solid line indicates the nominal process.  Fig. \ref{fig:GPSS_real} b) provides comparative detail of the distribution of the trajectories with respect to the constraints when the policy is deployed on the GP state space model (blue) and when deployed to the real uncertain process (red). As previously, the shaded regions indicate 99\% confidence intervals for process deviation and the solid line indicates the nominal process. It is observed that there is very little mismatch between the model and real process in this region of the state space and as a result the distributions of the first and second constraint ($g_1 \ \text{and} \ g_2$) are almost indistinguishable. Notably, however, there is indeed clear, but small amounts of mismatch between the nominal process trajectories on the GP state space model and the real process as demonstrated via the third constraint plot of $g_3$. Interestingly, this plot shows that 99\% of the  real process trajectories (the red region) are contained within the (blue) region described by the samples from the GP state space model. This indicates the potential that the offline GP model (epistemic) uncertainty, expressed via the variance of the posterior, could be able to provide constraint satisfaction and ensure safe RL policies.
\begin{table}[h!]
  \caption{Case Study: Comparison of probabilities of joint constraint satisfaction $F_{LB}(0)$ and $F_{SA}(0)$ and objective values of $\pi_C^*(\cdot, \theta)$ as learned via the methodology on the real process and GP state space model. The objective performance is quantified via the mean and variance due to process stochasticity. See Eq. \ref{eq:CSObj} for detail of the process objective.}
  \label{table:safeCCPO}
  \centering
  \begin{tabular}{llll}
    \toprule

    Process &$F_{LB}(0)$ & $F_{SA}(0)$ &  Process Objective (Eq. \ref{eq:CSObj}) \\
    \midrule
    Offline Gaussian process model & 1.0 & 1.0 & 15.29 +/- 0.11 \\
    Online real uncertain process  & 1.0 & 1.0 & 15.23 +/- 0.096  \\
    \bottomrule
  \end{tabular}
\end{table}
Table \ref{table:safeCCPO} demonstrates the utility of the algorithm in achieving constraint satisfaction as desired in both the offline model and real uncertain process. There is a small discrepancy between the performances of the two validations. This could be explained either due to the number of finite samples (500) used in assessment of policy performance, or via small amounts of nominal process mismatch between the GP model and the real process. If the latter view is taken and it is assumed the biomass and nitrate states are perfectly predicted (biomass is not included in the objective directly and nitrate is, but weakly), then this difference corresponds to a 0.375\% prediction error of the nominal lutein trajectory. In the following sections, the work detailed here is benchmarked against results observed from implementing control policies on the uncertain process determined via a) offline DO and b) NMPC. The approach to generation of these results is discussed in Section \ref{sec:Bench}.

\subsection{Comparison to Benchmark Methods}

The benchmark for this case study is provided by DO and NMPC, both of which are common approaches to process control. In the following sections the investigation provides comparative analysis to demonstrate the utility and limitations of the methodology. 

\subsubsection{Comparison to Dynamic Optimization}

In order to demonstrate the effects of process stochasticity for dynamic optimization (DO), control profiles were generated for the system (Eq. \ref{eq:LSYS}) from four different initial conditions, all of which are probable to be drawn from the initial state distribution detailed in Section \ref{sec:CS}.  As previously, all results are derived from 500 realisations of the real uncertain process model. The comparative performance of the DO benchmark is detailed by Table \ref{table:bench}. 
\begin{table}[h!]
  \caption{Case Study: Comparison of probabilities of joint constraint satisfaction $F_{LB}(0)$ and $F_{SA}(0)$ and objective values of $\pi_C^*(\cdot, \theta)$ under the proposed dynamic optimization (DO) benchmark. Four different results are reported for DO, corresponding to the four different initial conditions used to generate the control profile offline. The objective performance is quantified via the mean and variance due to process stochasticity. See Eq. \ref{eq:CSObj} for detail of the process objective.}
  \label{table:bench}
  \centering
  \begin{tabular}{lllll}
    \toprule

    Algorithm & Initial Conditions $\mathbf{x}_0$ &$F_{LB}(0)$ & $F_{SA}(0)$ &  Process Objective $J(\tau)$ \\
    \midrule
    DO I & [0.276, 784, 0.0] & 0.036 & 0.056 & 16.68 +/- 0.24 \\
    DO II & [0.273, 774, 0.0] & 0.046     & 0.068 & 16.68 +/- 0.25 \\
    DO III & [0.270, 765, 0.0] & 0.030   & 0.048 & 16.65 +/- 0.25 \\
    DO IV & [0.267, 755, 0.0] & 0.043 & 0.064 & 16.61 +/- 0.25 \\
    Proposed  &  $\mathbf{x}_0 \sim p(\mathbf{x}_0)$ & 1.0 & 1.0 & 15.23 +/- 0.096  \\
    \bottomrule
  \end{tabular}
\end{table}
From Table \ref{table:bench} it is clear that the effects of small amounts of stochasticity have dramatic implications for the probability of joint chance constraint satisfaction for DO. Both the statistically robust $F_{LB}(0)$ and the sample approximate $F_{SA}(0)$ are less than 0.06 for all DO control profiles. This highlights the utility of the method proposed in accounting for process stochasticity. It is also necessary to comment on the standard deviation of the performance with respect to the process objective as reported. The RL policy trained by the method achieves a lower variance in performance than that of the DO scheme. This is worth discussion as it highlights the ability of RL policies to naturally account for process stochasicity in a closed loop feedback control manner. Whereas, the variance of performance reported for the DO strategies is similar across all results and expresses the effects of process stochasticity on an open loop nominal (and deterministic) control policy.  However, it is also important to note that although the RL method proposed performs with respect to constraint satisfaction, it does not achieve as well as DO with respect to the expected unconstrained process objective $J(\tau)$. This is mainly because backing the nominal process away from the constraint boundaries in order to account for variability, will naturally incur a decrease in the nominal performance of the policy. However, as the process objective function is only reduced by 8\% and the constraints are satisfied with high probability, the current approach is still advantageous.

Despite the comparative benefits of RL, it is worth highlighting that the performance is sensitive to correct specification of initial state distribution, $p(\mathbf{x}_0)$, in offline training. Initialising the system in an initial state, $\mathbf{x}_0$, not well described by $p(\mathbf{x}_0)$ will likely lead to deterioration in the performance of the RL policy. Compared to the traditional NMPC approach in which process model can be continuously re-calibrated using online data, other advanced techniques \cite{wang2019incremental} could conceivably be applied given the slow dynamics under consideration in this case study.  Although updating RL online is out of current study’s scope, it is worth investigating in future work.

\subsubsection{Comparison to  Nonlinear Model Predictive Control}

\begin{figure}[h!]
    \centering
        \subfigure(a){%
            \includegraphics[scale=0.30]{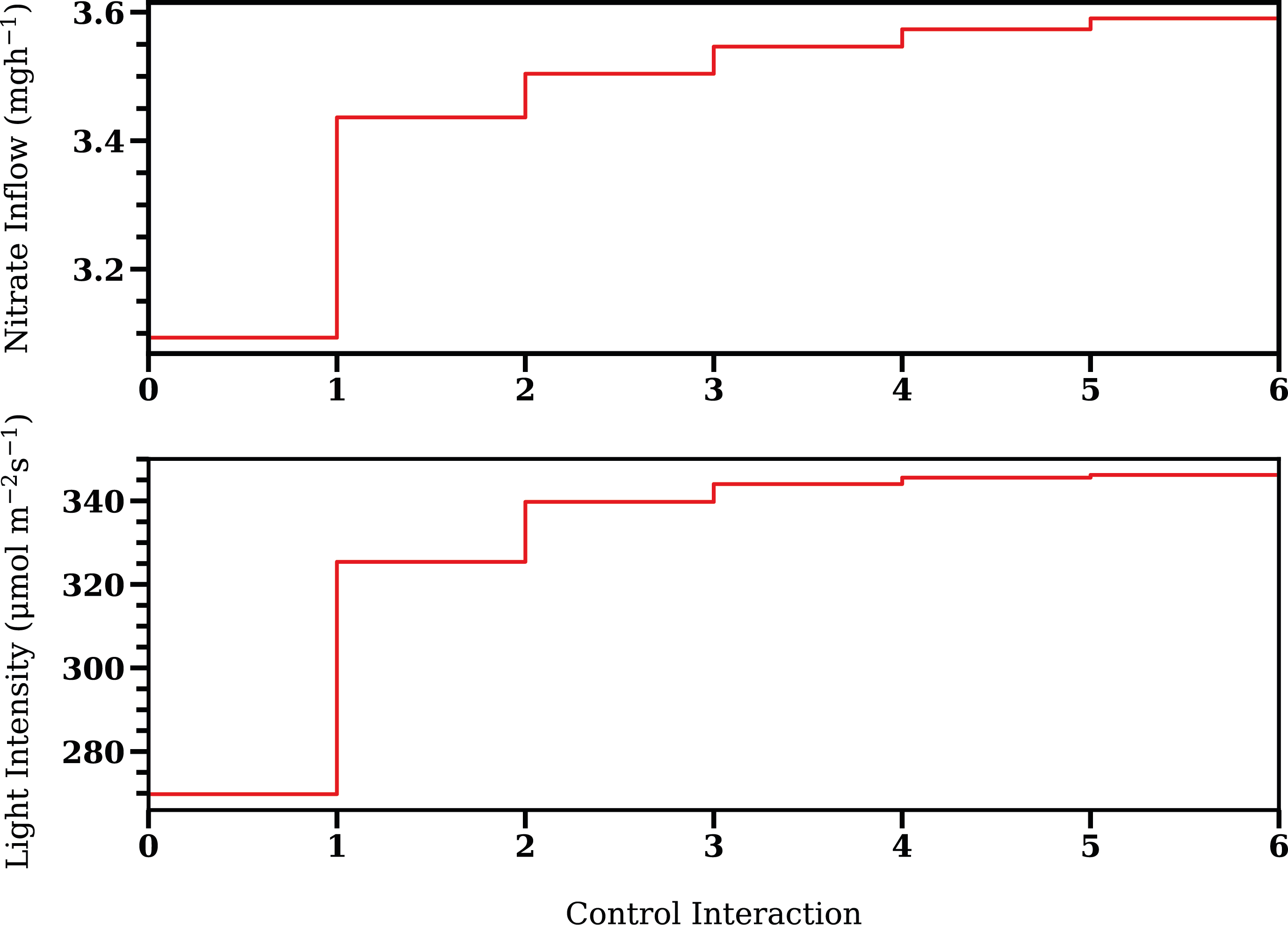}
            }
        \subfigure(b){%
            \includegraphics[scale= 0.30]{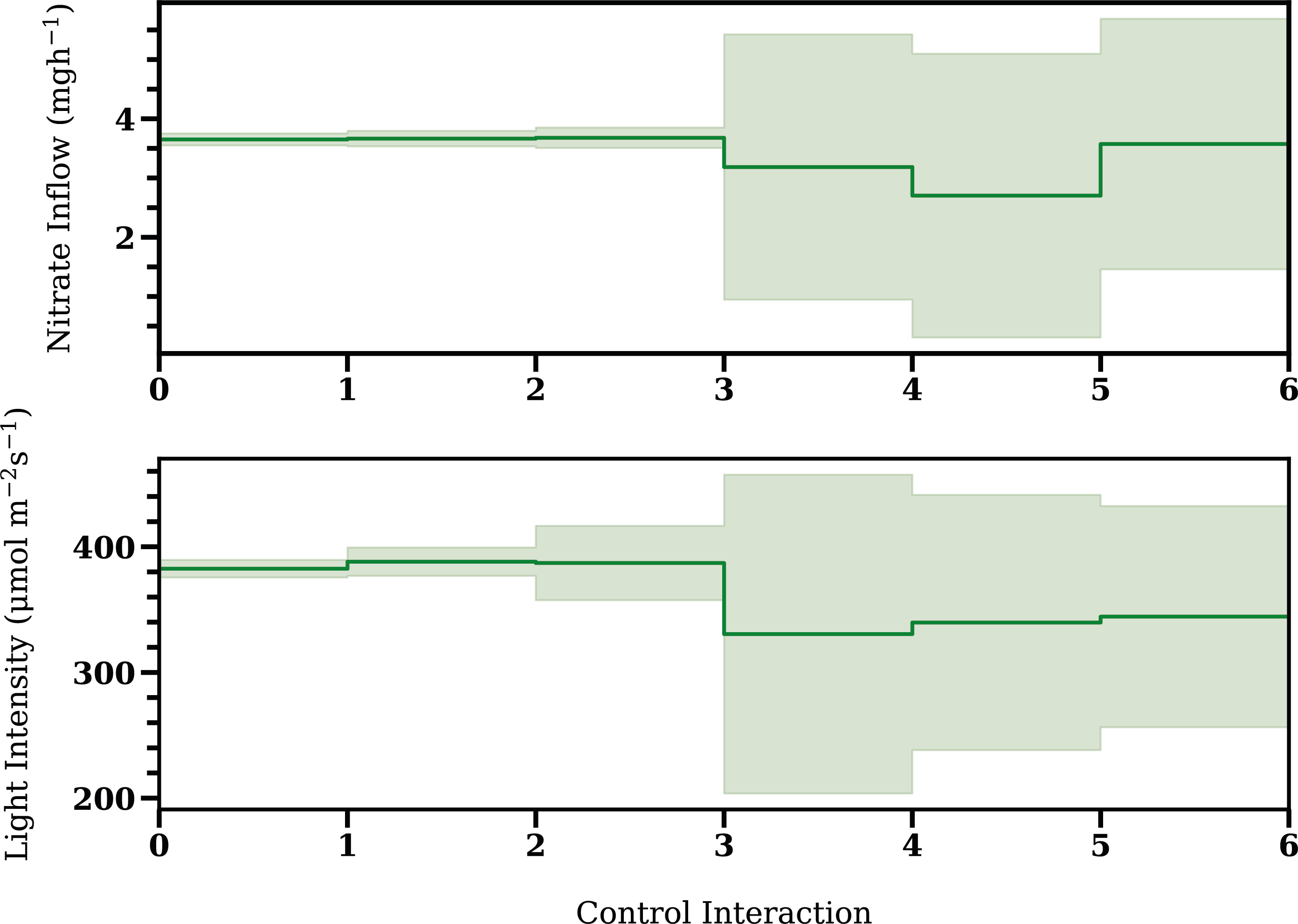}
            }
        \caption{Results from Case Study. (a) The distribution of controls selected by the RL policy, $\pi^*_C(\cdot, \theta)$, upon validation on the real uncertain process. Red solid line represents the average control trajectory and the light red shaded region represents a 1 standard deviation confidence interval (which is essentially non-existent), (b) The distribution of controls selected by the NMPC policy upon validation on the real uncertain process. Green solid line represents the average control trajectory and the light red shaded region represents a 1 standard deviation confidence interval}
        \label{fig:NMPCRLControls}
\end{figure}

The generation of the NMPC trajectory similarly assumes use of the deterministic variant of Eq. \ref{eq:LSYS}, as the process model. Here, however, the control policy is updated online via complete observation of the real uncertain process state (as is typical). The initial state is drawn from the initial state distribution detailed in Section \ref{sec:luteinCS}, which was also used to train and validate the RL policy $\pi_C^*(\cdot, \theta)$. Table \ref{table:bNMPC} reports the respective KPIs for the method proposed and the NMPC scheme.
\begin{table}[h!]
  \caption{Case Study: Comparison of probabilities of joint constraint satisfaction $F_{LB}(0)$ and $F_{SA}(0)$ and objective values of $\pi_C^*(\cdot, \theta)$ under the proposed benchmark of nonlinear model predictive control (NMPC). The objective performance is quantified via the mean and variance due to process stochasticity. See Eq. \ref{eq:CSObj} for detail of the process objective.}
  \label{table:bNMPC}
  \centering
  \begin{tabular}{llll}
    \toprule

    Algorithm &$F_{LB}(0)$ & $F_{SA}(0)$ &  Process Objective $J(\tau)$ \\
    \midrule
    NMPC  & 0.12 & 0.148 & 11.58 +/- 4.07  \\
    Proposed  & 1.0 & 1.0 & 15.23 +/- 0.096  \\
    \bottomrule
  \end{tabular}
\end{table}
Interestingly, with reference to Table \ref{table:bNMPC}, the method proposed performs better than NMPC with respect to the process objective. In this case, this is primarily due to the destabilisation of NMPC by process stochasitity, which was evidenced by the frequent inability to find control solutions online. This is common when stochastic systems are driven close to constraint boundaries with deterministic methods. The inability of NMPC to find control solutions online is the primary reason for the difference in objective performance as detailed by Table \ref{table:bNMPC} (Note: if solution could not be found, an approximate problem was solved to minimise constraint violation, and this was found to considerably improve performance - see Section \ref{sec:Bench} for information). In combination with worst cases of process stochasticity, this provides a skewing of the nominal process performance as reported. Demonstration of the sensitivity of the NMPC control scheme to process stochasticity is best expressed in analysis of the control trajectories generated in validation on the real uncertain process. This is reported by Fig. \ref{fig:NMPCRLControls}.

From Fig. \ref{fig:NMPCRLControls} the relative effect of stochasticity on the NMPC scheme is apparent. Fig. \ref{fig:NMPCRLControls} a) displays the distribution of controls selected under the RL policy, $\pi_C^*(\cdot, \theta)$, on the real uncertain process. The red solid line represents the average control trajectory and the red shaded region, which is essentially indistinguishable, represents a confidence interval of one standard deviation. Fig. \ref{fig:NMPCRLControls} b) represents the distribution of controls selected by the benchmark NMPC control policy when validated under the real uncertain process. Here, the green solid line represents the average control trajectory and the green shaded region, which is relatively large, represents a confidence interval of one standard deviation. It is likely that the average control trajectory plotted is not representative of actual control behaviour, i.e. the distribution of controls at each time interval is not best described by a unimodal Gaussian. However, the figure plotted well expresses the relative variance of controls selected.

From comparison of \ref{fig:NMPCRLControls} a) and b), it is clear that the RL method proposed naturally accounts for process stochasticity in a closed loop manner, with little variance in the distribution of controls shown. This is characteristic of a control strategy, which is robust to process uncertainty. This is especially beneficial in the context of cell cultivation or fermentation processes, where cell metabolism is sensitive to variation in the environmental conditions. As a result, the oscillatory control behaviour demonstrated by the NMPC control scheme would likely have a detrimental effect on the efficacy of operation/cell metabolism. It should be noted that the detrimental effects of process stochasticity on deterministic control strategies are demonstrated here, with small amounts of uncertainty. In the types of processes of concern to this work, (parametric) uncertainties can be much larger. This further contextualises the benefits provided by the strategy proposed i.e. the ability to simultaneously account for process-model mismatch and constraints.

The results provided by the work provokes the following question: is the primary benefit of the RL method proposed (relative to the NMPC result) derived due to the benefits of accounting for uncertainty in closed loop (i.e. operating within the MDP framework), or due to the description of process uncertainty provided by the GP model? Admittedly, it is difficult to answer this question certainly without further computational experiments and more thorough comparisons; however, through the current study it is believed that both elements are likely to be at play in separating the performance of the proposed method and NMPC. This question provides basis for future empirical studies.

\section{Conclusion}\label{sec:conc}

In this work, an efficient, purely data-driven method has been proposed, which considers the safe deployment of RL policies from the offline training environment (process model) to the real uncertain process. The method also provides approach to ensuring joint chance constraint satisfaction with a set probability. This is facilitated through use of the aleatoric and epistemic uncertainties expressed naturally by Gaussian process models, as well as the concept of constraint tightening. The method was analysed empirically and benchmarked against two commonly used, deterministic approaches to control and optimization of fed-batch process systems. It was demonstrated that the presence of even small amounts of process stochasticity may have a destabilising effect on the performance of deterministic methods and their relative probabilities of achieving joint constraint satisfaction.  It should be highlighted that the level of parametric uncertainty (5\%) expressed in this case study is a common lower-bound to that typically observed in the processes of concern to this work. It is likely that the benefits of this method would be even more apparent in cases where higher uncertainties were present. Therefore, it is thought that the scheme proposed is likely to be competitive with state-of-the-art sMPC approaches that similarly account for model uncertainties. The benefit (or drawback, depending on the context) of RL being that it shifts the computational effort offline, and is therefore much faster online (although slower offline). Further, the formalisation of this approach and the link drawn to reward shaping, enables combination of the method with any RL algorithm - policy optimization, action-value methods and all that lies inbetween. We hypothesise that once deployed, the policy could be continuously improved offline as the local model is iteratively improved and updated between batches \cite{rajeswaran2020game}. Further, it is possible the method could be adapted to the multi-agent setting for distributed control of fed-batch processes or into the domain of continuous processing \cite{mcclement2021metareinforcement}. We do however, assume the availability of an existing dataset, which provides information about the operational region of interest, however this could be developed using available mechanistic models and \cite{petsagkourakis2020safe} in a model-based design of experiments. Future work should consider the quantification of uncertainties in the parameterisation of the control function.

\bibliography{bibliography}
\appendix
\section{Appendices}
\subsection{Gaussian Process State Space Modelling}

\subsubsection{Training of Gaussian Process Models}\label{sec:GPtrain}
Selection of the appropriate hyperparameters, $\hat{\bm{\lambda}} = [\bm{\lambda}, \sigma^2_{n}] \in \mathbb{R}^{n_{\lambda} + 1}$ for a covariance function provides considerable improvement in the predictive abilities of GPs and can be viewed as a parallel to parameter estimation for mechanistic process models. The tuning procedure acts to maximise the marginal log-likelihood $p(\mathbf{Y}_j^T | \bm{\Upsilon}, \hat{\lambda})$ of the state specific, noisy output data points $\mathbf{Y}_j$, provided with the respective input measurements $\bm{\Upsilon}$ and hyperparameters $\hat{\lambda}$:

\begin{equation}\label{eq:maxlikelihood}
\begin{aligned}
    \log p(\mathbf{Y}_j^T | \bm{\Upsilon}, \hat{\lambda}) = -\frac{1}{2}(\mathbf{Y}_j(K + \sigma_{n}^2I_N)^{-1}\mathbf{Y}_j^T + \log |K + \sigma^2_{n}I_N| + N\log 2\pi)
\end{aligned}
\end{equation}

Gradient-based optimization may then be deployed to find $\hat{\lambda}$, which maximise the likelihood of observing our output data, given the covariance function chosen and the input data. This problem is non-convex and so typically multi-start schemes are deployed to find the best solution.

\subsubsection{Obtaining Function Realisations from GP State Space Models} \label{sec:MCGP}

In this work, we are concerned with obtaining function realisations from a Gaussian process state space model. The state space model is composed of $n_x$ individual Gaussian process models of each state. Here, we use the method proposed by \cite{strassen1969gaussian, bradford2020stochastic, umlauft2018scenario}. The method aims to update the posterior distribution of the Gaussian process model according to the initial dataset $\mathcal{D}$ used for model construction, as well as the states and control inputs observed during each trajectory evolution. This combined dataset is denoted $\mathcal{D}_+$. 

\vspace{0.2cm}
\begin{algorithm}[H]
%\SetAlgoLined
 \caption{Function Realisations via GP State Space Model for Decision-making Under Uncertainty}
\vspace{0.1cm}
%\hrule 
%\vspace{0.1cm}
\justify
\textbf{Initialise}: Experimental dataset $\mathcal{D}$; GP state space model $f_{GPSS} = [f^{1}_{GP}(\bm{\upsilon}), \ldots, f^{n_x}_{GP}(\bm{\upsilon})]$ with hyperparameters $\hat{\Lambda} = [\hat{\bm{\lambda}}_1, \ldots, \hat{\bm{\lambda}}_{n_x}]$ trained on $\mathcal{D} $; Control Policy $\pi(\mathbf{u}|\mathbf{x})$; Finite horizon length T; initial state distribution $p(\mathbf{x}_0)$; Memory for state $\mathcal{B}_x$ and control $\mathcal{B}_u$ trajectories, as well as for information related to decision making $\mathcal{B}_\pi$ for use in subsequent policy optimization. \vspace{0.1cm} \\ 
\textbf{1.} Set $\mathcal{D}_+ = \mathcal{D}$\vspace{0.1cm}\\
\textbf{2.} Draw $\mathbf{x}_0 \sim p(\mathbf{x}_0)$. Append $\mathbf{x}_0$ to $\mathcal{B}_x$\vspace{0.1cm}\\
\textbf{3.} \For{$t = 1, \ldots, T-1$}{

                \textbf{a.} Observe $\mathbf{x}_{t-1}$, sample $\mathbf{u}_{t-1} \sim \pi(\mathbf{u}|\mathbf{x})$ and concatenate, such that $\bm{\upsilon}_{t-1} = \big[ \mathbf{x}_{t-1}^T  \mathbf{u}_{t-1}^T \big]^T $\;
    
                \textbf{b.} Condition the GP state space model on ($\mathcal{D}_+, \bm{\upsilon}_{t-1}$) to obtain the predictive posterior:
                
                $ p(\mathbf{x}_{t} | \bm{\upsilon}_{t-1}, \mathcal{D}_+) = \mathcal{N}(\bm{\mu}(\bm{\upsilon}_{t-1}; \mathcal{D}_+), \bm{\Sigma}(\bm{\upsilon}_{t-1}; \mathcal{D}_+))$\;
    
                \textbf{c.} Draw next state from posterior of the GP state space model, $\mathbf{x}_t \sim p(\mathbf{x}_{t} | \bm{\upsilon}_{t-1}, \mathcal{D}_+)$ \vspace{0.1cm}\;
    
                \textbf{d.} Update $\mathcal{D}_+ = [\mathcal{D}^T_+ \ d_{N+t}^T]^T$, where $d_{N+t} = [\bm{\upsilon}^T_{t-1}\ \mathbf{x}^T_t]$ and append $\mathbf{x}_t$ and $\mathbf{u}_{t-1}$ to $\mathcal{B}_x$ 
                
                and $\mathcal{B}_u$, respectively\;
    }

\justify \textbf{Output}:  Function realisation stored in $\mathcal{B}_x$ and $\mathcal{B}_u$ and information related to decision making $\mathcal{B}_\pi$ (to be explained in \textbf{Algorithm 2})

\label{Alg:UncertProp}
\end{algorithm} \vspace{0.2cm}

The use of Algorithm \ref{Alg:UncertProp} allows for proper propagation of model uncertainty and sampling of functions from the GP. In essence, it is desired to obtain state sequences $\mathbf{x}_{0:T} = [\mathbf{x}_0, \ldots, \mathbf{x}_T]$, which are expressive of Eq. \ref{eq:nonlinDynA} and represent a realisation of process uncertainty. As samples $\mathbf{x}_{t}$ are drawn from the posterior they are added, along with the respective input $\bm{\upsilon}_{t-1}$, to the dataset $\mathcal{D}_+$ upon which the GP is conditioned. This leads to a subsequent update of the GP posterior distribution (via \ref{eq:A2} - \ref{eq:A3}) considering previous samples $\mathbf{x}_{t-1}$ as noiseless observations, with retention of the original covariance function hyperparameters $\bm{\hat{\lambda}}$. This means that if the updated GP posterior were to be queried at the previous input $\bm{\upsilon}_{t-1}$, the exact realisation of $\mathbf{x}_{t-1}$ would be drawn again i.e. the GP would express $\mathbf{x}_{t-1}$ deterministically. Such an outcome highlights the algorithm's utility in effective function space sampling and implies that future process evolution is explicitly dependent upon the past realisations of uncertainty.

We can express the updated posterior distribution of the ${j^{th}}$ GP after transition from one discrete time index at $t=t_0$ to $t=t_1$ as follows:

\begin{equation}\label{eq:A2}
    \begin{aligned}
        \mu_j(\bm{\upsilon}^*; \mathcal{D}_+) &= K_*^{+^T}\Sigma^{+^{-1}}\mathbf{Y}_j^{+^T} \\
        \sigma^2_j(\bm{\upsilon}^*; \mathcal{D}_+) &= k(\bm{\upsilon}^*, \bm{\upsilon}^*) - K_*^{+^T}\Sigma^{+^{-1}}K^+_*
    \end{aligned}
\end{equation}

where $\mathbf{Y}_j^+ = [\mathbf{Y}_j, y_j^+] \in \mathbb{R}^{1\times(N+1)}$, where $y_j^+\in \mathbb{R}$ is state $x_j\in \mathbb{R}$ observed at time index $t=t_1$; The updated covariance matrices are expressed as follows:

\begin{equation}\label{eq:A3}
    \begin{aligned}
        K_*^{+^T} = \big[K_*^T, k(\upsilon_*, \upsilon^+)\big] \qquad  \Sigma^+{^{-1}} = \bigg[ \begin{matrix} K + \sigma_n I_N & K_+ \\ K_+^T & k(\bm{\upsilon}^+, \bm{\upsilon}^+)  \end{matrix}  \bigg]^{-1}
    \end{aligned}
\end{equation}

where $K_+^T = [k(\bm{\upsilon}_+,\bm{\upsilon}_i), \ldots , k(\bm{\upsilon}_+,\bm{\upsilon}_N) ] \in \mathbb{R}^{1 \times N}$, and $\bm{\upsilon}_+ \in \mathbb{R}^{n_{\upsilon}}$ is the state and control input pair at time index $t=t_0$. This process is repeated iteratively for state transitions thereafter, which means that the memory and computation requirements will grow quadratically and cubically, respectively, with the time horizon $T$. Updating $K^+_*$, $Y^+_j$ is relatively easy, however, updating $\Sigma^{+^{-1}}$, is slightly more involved due to inversion. In order to do this we use the method from \cite{strassen1969gaussian} as proposed in \cite{bradford2020stochastic, umlauft2018scenario}. We refer the reader to these works for more information.

\subsection{Validation of Gaussian Process Models Used in Case Study}\label{sec:GPvalid}

Table \ref{table:validationGPSS} details the results of leave-one-out cross validation of the Gaussian process state space model used in this case study. Specifically, the results reported assess multi-step ahead predictions, which correspond to forecasting the entire batch given an initial state and control profile. Results are reported as the average across all possible different folds (of which there are 32). Predictions from the GP were drawn using the mean of the posterior. The dataset used to construct the Gaussian process models is available at https://github.com/mawbray/Lutein-Dataset
\begin{table}[h!]
  \caption{Multistep prediction mean absolute percentage error (MAPE) of leave-one-out cross validation of Gaussian process state space model used in Case study. }
  \label{table:validationGPSS}
  \centering
  \begin{tabular}{llll}
    \toprule

    Component of State & MAPE (\%) \\
    \midrule
    Biomass & 2.5 \\
    Nitrate  & 4.3\\
    Lutein & 2.2 \\
    \bottomrule
  \end{tabular}
\end{table}

\subsection{Proximal Policy Optimization, The Advantage Function and Entropy Regularisation} \label{sec:PPOdetails}
 PPO is at its core a policy gradient (PG) method. PG methods have previously been discussed, and so this work directs the interested reader to the original paper \cite{sutton1999policy} and other recent work \cite{petsagkourakis2019reinforcement}. PPO utilises a specific instance of the PG, known as the advantage policy gradient (APG). The APG \cite{mnih2016asynchronous} is a powerful, low variance form of the policy gradient, which utilises the generalised advantage function estimate $A_\varphi$ (GAE), rather than the action-value estimate, as in vanilla policy gradient methods \cite{sutton1999policy}. Further detail on the GAE and PPO is provided by \ref{sec:Adv} and \ref{sec:EntropyRegPPO}, respectively. In practice, the investigation found the addition of an entropy regularisation term useful in RL training. Entropy regularisation is widely studied in the RL literature, and at a high level provides mechanism to ensure the policy does not converge deterministically to a poor local optimum. This is particularly important in view of RL as a set of sampling-based algorithms \cite{neu2017unified, pmlr-v97-ahmed19a} and is discussed further in \ref{sec:entropyreg}.

\subsubsection{The Advantage Function}\label{sec:Adv}
The advantage function \cite{schulman2018highdimensional} is formalised:
\begin{equation}
    \begin{aligned}
        V^\pi(\mathbf{x}_t) = \mathbb{E}_\pi\bigg[\sum_{t'=t}^{T-1}R_{t'+1} | \mathbf{x} = \mathbf{x}_t\bigg] &\qquad Q^\pi(\mathbf{x}_t, \mathbf{u}_t) = \mathbb{E}_\pi\bigg[R_{t+1} + \gamma V^\pi(\mathbf{x}') | \mathbf{x} = \mathbf{x}_t, \mathbf{u} = \mathbf{u}_t\bigg]\\
        A^\pi(\mathbf{x}_t, \mathbf{u}_t) &= Q^\pi(\mathbf{x}_t, \mathbf{u}_t) - V^\pi(\mathbf{x}_t)\\
    \end{aligned}
\end{equation}

and, represents the difference between the expected returns under a policy in the current state, $V^\pi$, and the returns accumulated from selecting a given control in the current state and the current policy thereafter, $Q^\pi$. In RL practice, parameterisation of the value function $V_\psi$ is required in order to approximate the true value function $V^\pi$, such that $V_\psi \approx V^\pi$. Decision as to the model structure and initialisation of the parameters asserts bias into estimation of the advantage function. This is reduced through use of the generalised advantage function estimate $\hat{A}^\pi$ (GAE). The GAE provides a mechanism to explicitly \textit{trade off} variance and bias, by maximising the information provided by the reward signal. Explicitly, the GAE is formalised as:

\begin{equation}
    \begin{aligned}
        \hat{A}^\pi_{t} &=  \delta_{t+1} +(\rho \gamma) \delta_{t+2} +\ldots+(\rho \gamma)^{T-t+1} \delta_{T}\\
        \delta_{t+1} &= R_{t+1} + \gamma  V_\psi(\mathbf{x}_{t+1}) - V_\psi(\mathbf{x}_{t})
    \end{aligned}\label{eq:GAE}
\end{equation}

The parameter $\rho =[0,1]$ provides the mechanism to balance the bias and variance. Values closer to 1 reduce bias by utilising more information from the reward signal, but at the compromise of increasing the variance of the estimate. The opposite applies as values tend to 0. 

\subsubsection{Entropy Regularisation}\label{sec:entropyreg}

There is a rich literature on maximum entropy (Max.Ent.) RL \cite{ahmed2019understanding,haarnoja2018soft,ziebart2010modeling}. Instead of simply optimizing for the process objective and accumulated reward, $G(\bm{\tau})$, Max.Ent. RL also optimizes for the expected entropy of the stochastic policy learned. As a result, we can formulate the Max.Ent. RL objective, $J_H$ as follows:

\begin{equation}
    \begin{aligned}
        J_H = \mathbb{E}\big[G(\bm{\tau}) + H_\pi\big]
    \end{aligned}
\end{equation}

where $H_\pi = -\mathbb{E}_\pi\big[\log\pi(\mathbf{u}|\mathbf{x})\big]$ is the entropy of the policy. Typically, in practice, this objective is maximised via a regularisation term i.e. not as an extrinsic addition of entropy to the reward signal, and therefore not optimized via the PG. It is thought that entropy regularisation provides two main benefits: 1) it modifies the optimization landscape 'for the better', and in some cases provides a smoother landscape than the vanilla objective, and 2) the use of entropy plays some role in tackling the exploration-exploitation paradigm, i.e. by regularising entropy, exploration is encouraged, preventing convergence to a suboptimal deterministic policy. The use of entropy was found to be particularly helpful in this study, aiding the learning dynamics. It could be perceived that the constraint boundary provides a discontinuity in the reward landscape, and the promotion of exploration via entropy provides mechanism to 'escape' local optima.

\subsubsection{Entropy Regularised Proximal Policy Optimization}\label{sec:EntropyRegPPO}
PPO aims to provide conservative policy updates, by utilising the concept of trust region optimization. The idea of trust region optimization in the RL sense, is to constrain the update of an initial policy, such that the ultimate policy remains within a given distance of the initial in policy space. This distance could e.g. be quantified by the Kullback-Liebler divergence. One algorithm known as trust-region policy optimization (TRPO) necessitates estimate of the Hessian of the approximate KL divergence with respect to the policy parameters \cite{schulman2017trust} (this also shares similarities with the natural policy gradient \cite{kakade2001natural}). PPO sidesteps this complexity through approximation of the 2nd order TRPO update with a first order update - instead of explicitly enforcing this as a hard constraint, PPO enforces this via a penalty method \cite{schulman2017proximal}. This means that PPO is more computationally efficient than TRPO and provides flexible use of different function approximators (policy parameterisations). 

The objective function $L^{CLIP}$ formalised within the PPO framework follows:

\begin{equation}
    \begin{aligned}
        r_t(\theta) &= \frac{\pi_\theta(\mathbf{u}_t | \mathbf{x}_t)}{\pi_{\theta_{old}}(\mathbf{u}_t | \mathbf{x}_t)}\\
        L^{CLIP}(\theta) &= \hat{\mathbb{E}}_t\Big[\min(r_t(\theta){\hat{A}^\pi}_t, clip(r_t(\theta), 1-\epsilon, 1 + \epsilon){\hat{A}^\pi}_t)\Big]
    \end{aligned} \label{eq:PPOobj}
\end{equation}

where, $\epsilon = [0,1]$ and $\hat{A}^\pi_t$ is the advantage function, as discussed previously. By clipping the ratio $r$, updates corresponding to negative advantages are clipped with a ratio of $r=1+\epsilon$, whereas updates with positive advantages are clipped at $r=1-\epsilon$. The minimum is taken in order to provide a pessimistic update and enforce what could be interpreted as a trust-region. A full entropy regularised PPO algorithm is presented by Algorithm \ref{alg:EntPPO}.

% insert algorithm
\begin{algorithm}[H]
\SetAlgoLined
 \caption{Entropy Regularised Proximal Policy Optimization} \label{alg:EntPPO}
 \vspace{0.1cm}

\textbf{Initialise}: Approximate state space model or process dynamics $f_{SS}(\cdot)$; Initial control policy $\pi(\mathbf{u}|\mathbf{x}; \theta_0)$; Initial critic $V(\mathbf{x}_t, \psi_0)$; Reward function $R_{xx'}$; Finite horizon length $T$; initial state distribution $p(\mathbf{x}_0)$; entropy penalty term $\beta\in \mathbb{R}^+$; Learning rate $w^\pi \in \mathbb{R}^+$; Learning rate $w^V \in \mathbb{R}^+$; Strategies for updating the learning rates (schedules) $f_w^\pi(\cdot)$ and $f_w^V(\cdot)$; Memory $\mathcal{B}_{info}$ for information required for policy optimization; $K$ episodes; Learning updates per batch $J$; batchsize of $M$ trajectories; tolerance criterion\; \vspace{0.1cm}

\textbf{1.} $i = 0$\; \vspace{0.1cm}

\textbf{2.} \While{not converged}{\vspace{0.1cm}

                \justify \textbf{a.} Obtain a batch of $K$ rollouts over horizon of $T$ discrete intervals, via $\pi(\mathbf{u}|\mathbf{x}; \theta_i)$, $f_{SS}$, 
                
                and $p(\mathbf{x}_0)$\;
                
                \textbf{b.} Return trajectory information i.e. rewards ${R}^{(k)}_{0:T-1} = [{R}^{(k)}_1, \ldots, {R}^{(k)}_{T}]$ under $R_{xx'}$ 
                
                for the sequence of controls $\mathbf{u}^{(k)}_{0:T-1} = [\mathbf{u}^{(k)}_0, \ldots, \mathbf{u}^{(k)}_{T-1}]$ and states $\mathbf{x}^{(k)}_{0:T} = [\mathbf{x}^{(k)}_0, \ldots, \mathbf{x}^{(k)}_{T}]$, 
                
                corresponding to each rollout and store in $\mathcal{B}_{info}$\;\vspace{0.1cm}
                
                \textbf{c.} $j = 0$\;\vspace{0.1cm}
                
                    \While{$j < J$}{\vspace{0.1cm}
                    
                            \textbf{i.} Perform policy optimization by sampling the information of $M$ trajectories from 
                            
                            $\mathcal{B}_{info}$, calculating the respective importance ratios $r_t$ via Eq. \ref{eq:PPOobj} and GAEs via
                            
                            Eq. \ref{eq:GAE}:  \vspace{0.1cm}
                            
                            $\theta_{i+1} = \theta_{i} + w_{i} \nabla_\theta \big[\frac{1}{MT}\sum_{m=1}^{M}\sum_{t=0}^{T-1} L^{CLIP}(\mathbf{x}^{(m)}_t, \mathbf{u}^{(m)}_t,\mathbf{x}^{(m)}_{t+1},\theta_{i}) + \beta H_\pi(\pi(\mathbf{u}^{(m)}_t|\mathbf{x}^{(m)}_t))\big]$\;
                            
                            \textbf{ii.} Update the critic $V(\mathbf{x}, \psi_{i})$ on the same data sampled in \textbf{c.i.} and the respective 
                            
                            returns, $G_t$: \vspace{0.1cm} 
                            
                            $ \psi_{i+1} = \psi_i - \frac{1}{MT}\sum_{m=1}^M \sum_{t=0}^{T-1} \nabla_{\psi_i} V(\mathbf{x}, \psi_{i})(V(\mathbf{x}^{(m)}_t, \psi_{i}) - G^{(m)}_t)$ \vspace{0.1cm}\;
                            
                            \textbf{iii.} Update the learning rate : $ w^\pi_{i+1} = f_w^\pi(w_{i})$ \vspace{0.1cm}\; 
                            
                            \textbf{iv.} Update the learning rate : $ w^V_{i+1} = f_w^V(w_{i})$ \vspace{0.1cm}\;
                            
                            \textbf{v.} $i += 1$, $j+=1$ \vspace{0.1cm}\; }
                        
                \textbf{d.} Reset memory  $\mathcal{B}_{info}$\; 
    
                \textbf{e.} Assess tolerance criterion \vspace{0.1cm}\;}
%\textbf{Output:} Optimal Constrained Policy $\pi_C^*(\theta)$ trained under $\bm{\xi}^*$    
\textbf{Output:}  Optimal policy $\pi(\theta^*)$ and critic $V(\psi^*)$\;
\end{algorithm} 

\subsection{Evaluating Joint Constraint Satisfaction Empirically} \label{sec:jointSA}
In this work, we are concerned with the satisfaction of the joint chance constraints expressed by:

\begin{equation}
    \begin{aligned}
        F_X(0) &= \mathbb{P}(X \leq 0) = \mathbb{P}(\bigcap_{i=0}^T \{\textbf{x}_i \in \mathbb{\hat{X}}_i\})
    \end{aligned}\label{eq:jointprobcdfAppend}
\end{equation}

where $\mathbb{\hat{X}}_i$ is the tightened joint constraint set and $${X} = \max_{(t,j) \in \{0, \ldots, T\} \times \{1, \ldots, n_g\}} A_j\mathbf{x}_{t} - b_j, $$

defines the maximum constraint violation during process evolution. As analytical expression of Eq. \ref{eq:jointprobcdfAppend} is not available, it is proposed to instead estimate it via Monte Carlo sampling. Hence we can define the empirical cumulative distribution function (ecdf) via $S$ Monte Carlo samples:

\begin{equation}
    F_X(0) \approx F_{SA}(0) = \frac{1}{S}\sum_{s=1}^S  \mathbbm{1}(X^{(s)} \leq 0)
\end{equation}

where $\mathbbm{1}$ is the indicator function. However, due to the limits imposed by finite samples, the approximation is likely to include error. Therefore, in order to account for this we deploy a concept from the \textit{binomial proportion confidence interval} literature. Specifically, the Clopper–Pearson interval \cite{clopperpearson}, which enables us to ensure the probability of joint satisfaction with a given confidence level, $1-\upsilon$, on the basis of empirical observation. This is expressed by Lemma \ref{lemma:PC}, which is recycled from \cite{petsagkourakis2020chance}. 

\begin{lemma}\label{lemma:PC} 
\textbf{Joint chance constraint satisfaction via the Clopper-Pearson confidence interval} \cite{clopperpearson,Intervalestimation}: Consider the realisation of $F_{SA}(0)$ based on $S$ independently and identically distributed samples. The lower bound of the true value $F_{LB}(0)$ may be defined with a given confidence $1-\upsilon$, such that:

\begin{equation}
    \begin{aligned}
        &\mathbb{P}(F_X(0) \geq F_{LB}(0)) \geq 1 - \upsilon\\
        &F_{LB}(0) = 1 - \text{betainv}(\upsilon, S+1-SF_{SA}(0), SF_{SA}(0)) 
    \end{aligned}
\end{equation}
where $\text{betainv}(\cdot)$ is the inverse of the Beta cumulative distribution function with parameters $\{S+1-SF_{SA}(0)\}$ and $\{SF_{SA}(0)\}$.

\end{lemma}

\subsection{Further Information on Benchmark}\label{sec:benchmod}

In construction of the benchmark provided, a direct collocation scheme was implemented in python. The code is available at \text{https://github.com/mawbray/Lutein\_DO}. In the case of NMPC and online optimization, an approximate problem was solved if a solution could not be found online an approximate problem was solved instead. This was conducted via the following formulation: 
\begin{equation}
\begin{aligned}
    \max_{\mathbf{u}_{t{'}:T-1}} &\sum_{t=t{'}}^{T-1} R_{t+1} - \mathbf{z}(A\mathbf{x}_t -b)   \quad \quad \quad \quad  (\text{see Eqs. \ref{eq:CSObj} and \ref{eq:cons}})\\
    \text{s.t.}&\\
    &\textbf{x}_{t+1} = f(\textbf{x}_t, \textbf{u}_t) \quad \quad \quad \quad \quad \quad \quad (\text{see Eqs. \ref{eq:LSYS} and \ref{eq:kinetics}})\\
    &\textbf{u}_t\in\mathbb{\hat{U}}\\
    & A\textbf{x}_t - b \leq 0 \quad \quad \quad \quad \quad \quad \quad \quad \quad (\text{see Eqs. \ref{eq:safeset} and \ref{eq:cons}})\\
    &  \forall t \in \left\{t{'},...,T-1\right\}\label{eq:modOCP}
\end{aligned}
\end{equation}
where $\mathbf{z} = [1, 10, 10]$ and $\mathbf{x}_{t{'}}$ is observed from the uncertain process. This approximate problem modifies the objective function to incentivize minimisation of constraint violation. It should be stressed that this problem is only solved if a solution cannot be found to the original problem. This was typically the case when the optimization was initialised such that $\mathbf{x}_{t{'}}$ had already violated the constraints and arose from the inability to handle constraints.

\subsection{Hyperparameters for Learning in Case Study}
\begin{table}[h!]
    \caption{Miscellaneous hyperparameters specific to Proximal policy optimization algorithm used in this work.}
    \label{table:safeCCPO_params}
    \centering
    \begin{tabular}{l|l}
        \toprule
         Parameter & Value  \\
         \midrule
         Episodes, $K$ & 200 \\
         Nodes per LSTM layer of Policy Net. & 30\\
         LSTM Layers in Policy Net. & 4 \\
         Activation function in output layer of Policy Net. & ReLU6 \\
         Nodes per LSTM layer in Value Net. & 30\\
         LSTM layers in Value Net. & 2 \\
         Activation function in output layer of Value Net. & Leaky ReLU \\
         Policy learning rate, $w^\pi$ & $5 \times 10^{-3}$\\
         Value learning rate, $w^V$ & $5 \times 10^{-3}$\\
         GAE weight, $\rho$ & $0.99 $\\
         Batch size, $M$ & $100 $\\
         Weight updates, $J$ & $2$ \\
         Clipping factor, $\epsilon$ & $0.2$\\
         Discount factor, $\gamma$ & $0.99$\\
         Entropy regularisation weights, $\beta$ & $5 \times 10^{-2}$\\
         \bottomrule
    \end{tabular}
\end{table}
\end{document}